\documentclass{article}

\usepackage{times}
\usepackage{graphicx} % more modern
\usepackage{subfigure}

\usepackage{natbib}

\usepackage{algorithm}
\usepackage{algorithmic}

\usepackage[draft]{hyperref}

\usepackage{url}
\usepackage{graphicx}
\usepackage{wrapfig}
\usepackage{amsmath,amssymb,amsthm}

\usepackage[accepted]{icml2016}

\usepackage[algo2e, ruled, noline]{algorithm2e}

\providecommand{\DontPrintSemicolon}{\dontprintsemicolon}
\DontPrintSemicolon

\icmltitlerunning{Dueling Network Architectures for Deep Reinforcement Learning}

\newcommand{\E}{\mathbb{E}}

\newcommand{\expect}[1]{\mathbb{E}\left[ {#1} \right]}
\newcommand{\expectQ}[2]{\mathbb{E}_{{#2}} \left[ {#1} \right]}

\newcommand{\be}{\begin{equation}}
\newcommand{\ee}{\end{equation}}
\newcommand{\bea}{\begin{eqnarray}}
\newcommand{\eea}{\end{eqnarray}}
\newcommand{\beaa}{\begin{eqnarray*}}
\newcommand{\eeaa}{\end{eqnarray*}}

\DeclareMathAlphabet{\mathpzc}{OT1}{pzc}{m}{n}
\DeclareMathOperator*{\argmax}{arg\,max}

\begin{document}

\twocolumn[
\icmltitle{Dueling Network Architectures for Deep Reinforcement Learning}

% from ICLR
%\author{Ziyu Wang, Nando de Freitas \& Marc Lanctot\\
% \thanks{ Use footnote for providing further information
% about author (webpage, alternative address)---\emph{not} for acknowledging
% funding agencies.  Funding acknowledgements go at the end of the paper.} \\
%Google DeepMind\\
%London, UK\\
%\texttt{\{ziyu,nandodefreitas,lanctot\}@google.com} \\
%}

% It is OKAY to include author information, even for blind
% submissions: the style file will automatically remove it for you
% unless you've provided the [accepted] option to the icml2016
% package.
\icmlauthor{Ziyu Wang}{ziyu@google.com}
% \icmladdress{Google DeepMind, London, UK}
\icmlauthor{Tom Schaul}{schaul@google.com}
% \icmladdress{Google DeepMind, London, UK}
\icmlauthor{Matteo Hessel}{mtthss@google.com}
% \icmladdress{Google DeepMind, London, UK}
\icmlauthor{Hado van Hasselt}{hado@google.com}
% \icmladdress{Google DeepMind, London, UK}
\icmlauthor{Marc Lanctot}{lanctot@google.com}
% \icmladdress{Google DeepMind, London, UK}
\icmlauthor{Nando de Freitas}{nandodefreitas@gmail.com}
\icmladdress{Google DeepMind, London, UK}

% You may provide any keywords that you
% find helpful for describing your paper; these are used to populate
% the "keywords" metadata in the PDF but will not be shown in the document
\icmlkeywords{}

\vskip 0.3in
]

\begin{abstract}
In recent years there have been many successes of using deep representations in reinforcement learning. Still, many of these applications use conventional architectures, such as convolutional networks, LSTMs, or auto-encoders. In this paper, we present a new neural network architecture for model-free reinforcement learning. Our dueling network represents two separate estimators: one for the state value function and one for the state-dependent action advantage function. The main benefit of this factoring is to generalize learning across actions without imposing any change to the underlying reinforcement learning algorithm. Our results show that this architecture leads to better policy evaluation in the presence of many similar-valued actions. Moreover, the dueling architecture enables our RL agent to outperform the state-of-the-art on the Atari 2600 domain. 
%Double DQN method of~\citet{vanHasselt:2015} in 46 out of 57 Atari games.
\end{abstract}

\section{Introduction}
\label{sec:introduction}

%!TEX root = main.tex

Over the past years, deep learning has contributed to dramatic advances in scalability and performance of machine learning~\cite{LeCun:2015}. One exciting application is the sequential decision-making setting of reinforcement learning (RL) and control. Notable examples include deep Q-learning \cite{Mnih:2015}, deep visuomotor policies \cite{levine2015end}, attention with recurrent networks \cite{Ba:2015}, and model predictive control with embeddings \cite{Watter:2015}. Other  recent successes include massively parallel frameworks~\cite{Nair:2015} and expert move prediction in the game of Go~\cite{Maddison:2015}, which produced policies matching those of Monte Carlo tree search programs, and squarely beaten a professional player when combined with search~\cite{alphago}.

\begin{figure}[b!]
\begin{center}
	\includegraphics[scale=0.3]{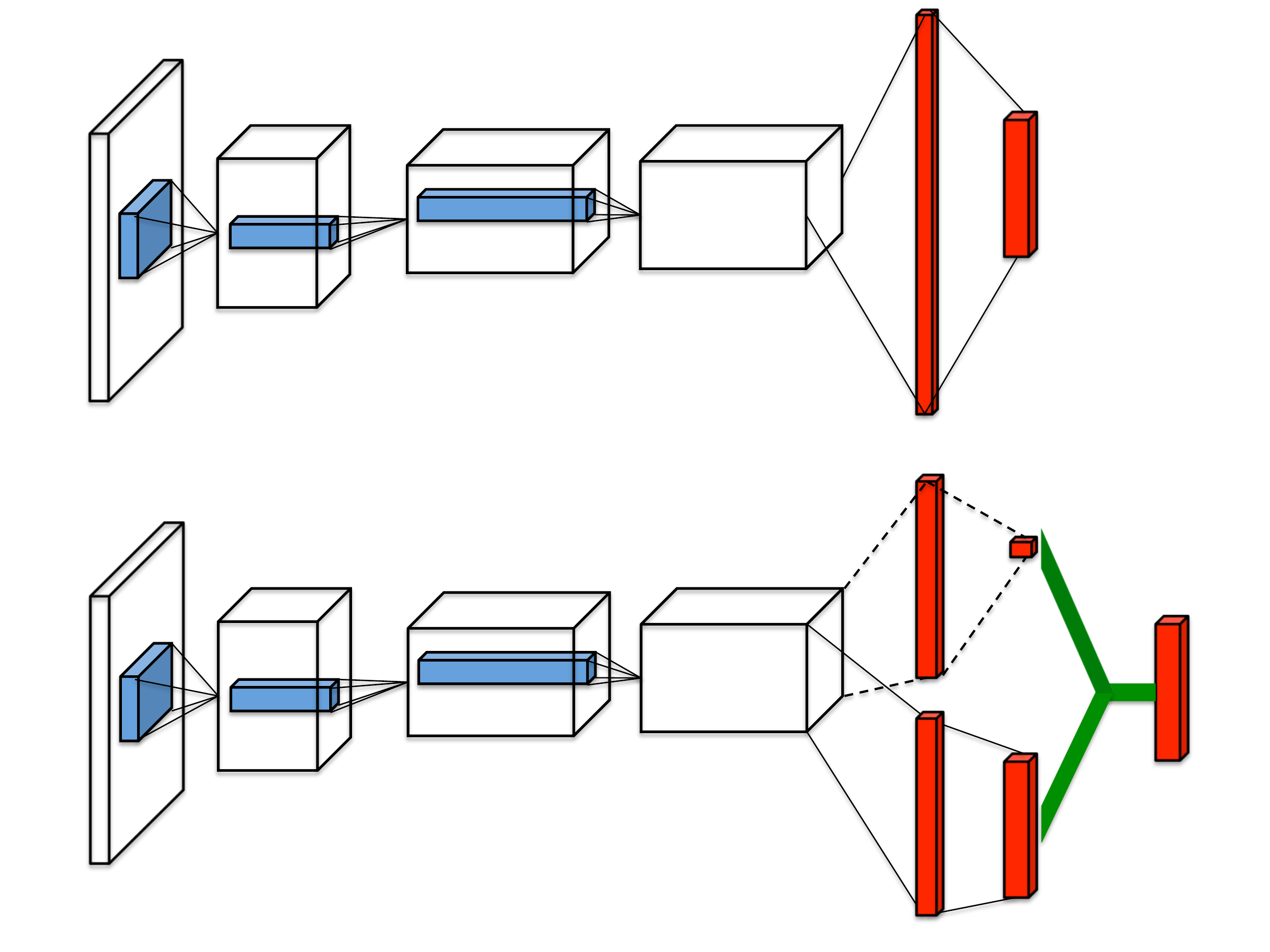}
\end{center}
\vspace{-5mm}
\caption{A popular single stream $Q$-network ({\bf top}) and the dueling $Q$-network ({\bf bottom}). The dueling network has two streams to separately estimate (scalar) state-value and the advantages for each action; the green output module implements equation (\ref{eq:combo2}) to combine them. Both networks output $Q$-values for each action.}
\label{fig:duelnet}
\end{figure}

In spite of this, most of the approaches for RL use standard neural networks, such as convolutional networks, MLPs, LSTMs and autoencoders. The focus in these recent advances has been on designing improved control and RL algorithms, or simply on incorporating existing neural network architectures into RL methods. Here, we take an \emph{alternative but complementary approach} of focusing primarily on innovating a neural network architecture that is better suited for model-free RL. This approach has the benefit that the new network can be easily combined with existing and future algorithms for RL. That is, this paper advances a new network (Figure~\ref{fig:duelnet}), but uses already published algorithms. 
%\begin{wrapfigure}{t!}{0.3\textwidth} 
%\vspace{-9pt}
%\vspace{-20pt}
%\end{wrapfigure}

The proposed network architecture, which we name the {\it dueling architecture}, explicitly separates the representation of state values and (state-dependent) action advantages. The dueling architecture consists of two streams that represent the value and advantage functions, while sharing a common convolutional feature learning module. The two streams are combined via a special aggregating layer to produce an estimate of the state-action value function $Q$ as shown in Figure~\ref{fig:duelnet}. This dueling network should be understood as a single $Q$ network with two streams that replaces the popular single-stream $Q$ network in existing algorithms such as Deep Q-Networks \citep[DQN;][]{Mnih:2015}. The dueling network automatically produces separate estimates of the state value function and advantage function, without any extra supervision. 
%The new special aggregator that makes this possible is described in Section 3.

\begin{figure}[t!]
\begin{center}
{\sc \small \hspace{0.6cm} Value \hspace{1.8cm} Advantage } \\
	\includegraphics[scale=0.55]{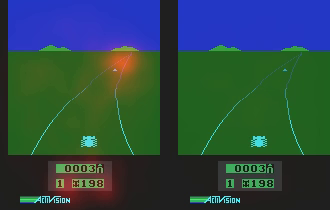} \\
{\sc \small \hspace{0.6cm} Value \hspace{1.8cm} Advantage } \\
	\includegraphics[scale=0.55]{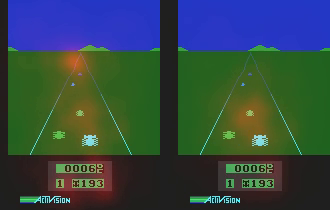}
\end{center}
\caption{See, attend and drive: Value and advantage saliency maps (red-tinted overlay) on the Atari game Enduro, for a trained dueling architecture. The value stream learns to pay attention to the road. The advantage stream learns to pay attention only when there are cars immediately in front, so as to avoid collisions.}  
\label{fig:saliency}
\end{figure}

Intuitively, the dueling architecture can learn which states are (or are not) valuable, without having to learn the effect of each action for each state. This is particularly useful %when the agent encounters itself 
in states where its actions do not affect the environment in any relevant way. To illustrate this, consider the saliency maps shown in Figure~\ref{fig:saliency}\footnote{\url{https://www.youtube.com/playlist?list=PLVFXyCSfS2Pau0gBh0mwTxDmutywWyFBP}}.
These maps were generated by computing the Jacobians of the trained value and advantage streams with respect to the input video, following the method proposed by~\citet{Simonyan:2013}. (The experimental section describes this methodology in more detail.) The figure shows the value and advantage saliency maps for two different time steps. In one time step (leftmost pair of images), we see that the value network stream pays attention to the road and in particular to the horizon, where new cars appear. It also pays attention to the score. The advantage stream on the other hand does not pay much attention to the visual input because its action choice is practically irrelevant when there are no cars in front. However, in the second time step (rightmost pair of images) the advantage stream pays attention as there is a car immediately in front, making its choice of action very relevant. 

In the experiments, we demonstrate that the dueling architecture can more quickly identify the correct action during policy evaluation as redundant or similar actions are added to the learning problem. 

We also evaluate the gains brought in by the dueling architecture on the challenging Atari 2600 testbed. Here, an RL agent with the same structure and hyper-parameters must be able to play 57 different games by observing image pixels and game scores only. The results illustrate vast improvements over the single-stream baselines of~\citet{Mnih:2015} and~\citet{vanHasselt:2015}. The combination of
prioritized replay~\citep{Schaul:2015} with the proposed dueling network results in the new state-of-the-art for this popular domain.

%\begin{figure}[h!]
%\begin{center}
%	\includegraphics[scale=0.55]{./figs/nature_comp}
%\end{center}
%\caption{\label{fig:natureComp}Comparison against DQN \cite{mnih2015human}
%in human performance percentage. The gray dashed horizontal line marks human level 
%performance (75\% of expert score). The Dueling architecture beats DQN in 49 out of 57 games. 
%}
%\end{figure}

% By computing the Jacobians for the value and advantage branches of the dueling architecture, we obtain saliency videos illustrating which aspects of the input video stream the model thinks are valuable and which it thinks it must attend to when immediate action is needed. 
%  

%TODO(@Marc): MORE EXAMPLES OF RECENT DEEP LEARNING SUCCESSES + REFERENCES TO EXISTING ONES

% Adding a related work section in the short term for volume :)

\subsection{Related Work}

The notion of maintaining separate value and advantage functions goes back to \citet{Baird:1993}. In Baird's original advantage updating algorithm, the shared Bellman residual update equation is decomposed into two updates: one for a state value function, and one for its associated advantage function. Advantage updating was shown to converge faster than Q-learning in simple continuous time domains in~\cite{Harmon:1995}. Its successor, the advantage learning algorithm, represents only a single advantage function~\cite{Harmon:1996}.
%One of the goals of advantage learning is to increase the gap between the optimal action and suboptimal actions \cite{Baird:1999}.

% There is evidence that these Bellman residual algorithms can help stabilize when learning with gradient-based function approximation~\cite{Baird:1995,Baird:1999}. A very recent contribution is the consistent Bellman operator~\cite{Bellemare2015Persistent}, which also showed improvements in the Atari domain. We believe this improvement is independent and would be complementary, and we leave it as an interesting direction for future work.

%Actually, I don't know that citing this one makes much sense...
%Advantage functions have also been used when learning parsers for structured prediction using inverse reinforcement learning~\cite{Neu:2009}.

The dueling architecture represents both the value $V(s)$ and advantage $A(s,a)$ functions with a single deep model whose output combines the two to produce a state-action value $Q(s,a)$. Unlike in advantage updating, the representation and algorithm are decoupled by construction. Consequently, the dueling architecture can be used in combination with a myriad of model free RL algorithms. 
%Moreover, the representation of value and advantage is very rich, incorporating shared deep feature learning modules.

%Unlike advantage updating, it does not require two separate update rules. Similar to advantage learning, the dueling network computes a single state-action value. However, w
%We are unaware of any work that attempts to model these functions {with a shared deep representation}.
%, and whose updates are controlled by common gradient.  

% While learning is still model-free as in deep Q-learning~\cite{Mnih:2015}, the network's internal representation of $V^{\pi}(s)$ acts as a simple local model for each state's policy value, as experience replay does on the larger scale~\cite{vanSeijen:2015}. This local value learning can then simplify learning $Q^{\pi}(s,a)$ values across similar actions in state $s$ by focusing on the relative differences for each action $a$ from $s$.

There is a long history of advantage functions in policy gradients, starting with \cite{Sutton:2000}.
As a recent example of this line of work, ~\citet{schulman2015advantage} estimate advantage values online to reduce the variance of policy gradient algorithms.

There have been several attempts at playing Atari with deep reinforcement learning, including \citet{Mnih:2015,Guo:2014,Stadie:2015,Nair:2015,vanHasselt:2015,Bellemare2015Persistent} and \citet{Schaul:2015}. The results of \citet{Schaul:2015} are the current published state-of-the-art.

%TODO(@Marc): ADD REF TO SCHULMAN'S ADVANTAGE ESTIMATION + PARSING/IRL PAPER

%modeling these values within the representation could lead to better generalization across similar actions and lessen the burden of engineering algorithmic improvements. Furthermore, this construction could allow more complex (e.g. contextual) interactions to be learned between state and action values.

\section{Background}
\label{sec:DRL}

%!TEX root = main.tex

% In this section, we formalize the terminology and notation used in the rest of the paper. We present only a brief summary. For an introduction we refer the reader to~\citep{SuttonBarto:1998}.

We consider a sequential decision making setup, in which 
an agent interacts with an environment $\mathcal{E}$ over discrete time steps, see~\citet{SuttonBarto:1998} for an introduction. In the Atari domain, for example, the agent perceives a video $s_t$ consisting of $M$ image frames:  $s_t =(x_{t-M+1}, \ldots, x_t) \in \mathcal{S}$ at time step $t$. The agent then chooses an action from a discrete set $a_t \in \mathcal{A} = \{1,\ldots,|\mathcal{A}|\} $ and observes a reward signal $r_t$ produced by the game emulator. 
%We represent $\mathcal{E}(s_t,a_t)$ as a random variable whose outcome is the next frame $x_{t+1}$ yielding the successor state $s_{t+1} \in \mathcal{S}$.

%The agent seeks to maximize the  discounted return
%is defined as $R_t = \sum_{\tau=t} \gamma^{\tau-t} r_{\tau}$,
%where $\gamma \in [0,1]$ is a discount factor that trades-off the importance of immediate and future rewards.
%In our experiments we use finite-length {\it episodes} but here we present the general case of continuing tasks.
%A policy $\pi : \mathcal{S} \rightarrow \Delta(\mathcal{A})$ is a function that maps each state to a probability distribution over actions.
The agent seeks maximize the expected discounted return, where we define the discounted return as $R_t = \sum_{\tau=t}^{\infty} \gamma^{\tau-t} r_{\tau}$. In this formulation,
$\gamma \in [0,1]$ is a discount factor that trades-off the importance of immediate and future rewards.  

For an agent behaving according to a stochastic policy $\pi$, the values of the state-action pair $(s,a)$ and the state $s$ 
% at time $t$ 
are defined as follows
\begin{eqnarray}
Q^{\pi}(s,a) &=& \expect{\left. R_t \right| s_t = s, a_t = a, \pi},~\mbox{ and } \nonumber \\
V^{\pi}(s) &=& \expectQ{Q^{\pi}(s,a)}{a \sim \pi(s)}.
\end{eqnarray}

The preceding state-action value function ($Q$ function for short) can be computed recursively with dynamic programming:
\[
Q^{\pi}(s,a)  = \expectQ{r + \gamma \expectQ{Q^{\pi}(s',a')}{a'\sim\pi(s')}~|~s,a,\pi}{s'}.
\]
We define the optimal $Q^{*}(s,a) = \max_{\pi} Q^{\pi}(s,a).$
Under the deterministic policy $a = \argmax_{a' \in \mathcal{A}} Q^{*}(s,a')$, it follows that
$V^{*}(s) = \max_{a} Q^{*}(s,a).$ From this, it also follows that the optimal $Q$ function satisfies the Bellman equation:
\be
Q^{*}(s,a) = \E_{s'} \left[ r +  \gamma  \max_{a' } Q^{*}(s',a')~|~s,a \right]. 
\ee

We define another important quantity, the {\it advantage function}, relating the value and $Q$ functions:
\be
\label{eq:advantage}
A^{\pi}(s,a) = Q^{\pi}(s,a) - V^{\pi}(s).
\ee
Note that $\expectQ{A^{\pi}(s,a)}{a \sim \pi(s)} = 0$. Intuitively, the value function $V$ measures the how good it is to be in a particular state $s$. The $Q$ function, however, measures the the value of choosing a particular action when in this state. The advantage function subtracts the value of the state from the Q function to obtain a relative measure of the importance of each action.

%TODO(@Marc): IMPROVE INTUITIONS. ADD MORE ON ADVANTAGE.
% In many domains, for some fixed model capacity (i.e. number of weights in the network) and number of training episodes, learning a $Q$ function is often more difficult than learning a value function. This is because the $Q$ function must represent both the state {\it and} advantage value per action in a single function, so models need to condition each action's value on each state $s \in \mathcal{S}$. In contrast, value functions simply need to output a single numerical value. Also, this increase in difficulty rises as $|\cal{A}|$ grows, which is problematic for RL on complex domains.

% However, the choice of whether learn a model that represents $V$ instead of $Q$ is not so simple, because policies are not directly obtainable from value functions alone. A deterministic policy based on a value function would need access to a model of the environment to be able to compute $\max_{a \in \mathcal{A}} \E_{s' \sim \mathcal{E}} [ V(s')~|~s,a]$, which is often prohibitively expensive or impossible in practice. The dueling architecture described in Section~\ref{sec:dueling} remains model-free using a $Q$ function, but which is decomposed into $V$ and $A$ to alleviate some of the difficulty caused by state-action conditioning.

\subsection{Deep Q-networks}
\label{sec:dqn}

The value functions as described in the preceding section are high dimensional objects. To approximate them, we can use a deep $Q$-network: $Q(s,a;\theta)$ with parameters $\theta$. To estimate this network, we optimize the following sequence of loss functions at iteration $i$:
\be
L_i(\theta_i) = \E_{s,a,r,s'} \left[ \left( y_i^{DQN} - Q(s,a; \theta_{i}) \right)^{2} \right],
\label{eq:loss}
\ee
with 
\be
y_i^{DQN} =  r + \gamma \max_{a'} Q(s',a';\theta^{-}),
\ee
where $\theta^{-}$ represents the parameters of a fixed and separate {\it target network}. % , which has shown to significantly increase the stability of the learning~\cite{Mnih:2015,vanHasselt:2015}.
We could attempt to use standard $Q$-learning to learn the parameters of the network $Q(s,a; \theta)$ online. However, this estimator performs poorly in practice. A key innovation in \citep{Mnih:2015} was to freeze the  parameters of the target network  $Q(s',a';\theta^{-})$ for a fixed number of iterations while updating the {\it online network} $Q(s,a; \theta_{i})$ by gradient descent. (This greatly improves the stability of the algorithm.) The specific gradient update is 
\begin{eqnarray*}
\hspace{-1.5em}&\nabla_{\theta_i} L_i(\theta_i) =
\displaystyle{\E_{s,a,r,s'}} \Big[ \hspace{-1mm}\left( y_i^{DQN} - Q(s,a; \theta_{i}) \right)\hspace{-1mm} \nabla_{\theta_i} Q(s,a; \theta_{i}) \Big]
\nonumber&
\end{eqnarray*}

This approach is model free in the sense that the states and rewards are produced by the environment. It is also off-policy because these states and rewards are obtained with a behavior policy (epsilon greedy in DQN) different from the online policy that is being learned.

Another key ingredient behind the success of DQN is  {\it experience replay}~\citep{Lin:1993,Mnih:2015}. During learning, the agent accumulates a dataset $\mathcal{D}_t =\{e_1, e_2, \ldots, e_t\}$ of experiences
$e_t=(s_t, a_t, r_t, s_{t+1})$ from many episodes. When training the $Q$-network, instead only using the current experience as prescribed by standard temporal-difference learning, the network is trained by sampling mini-batches of experiences  from $\mathcal{D}$ uniformly at random. The sequence of losses thus takes the form
\be
L_i(\theta_i) = \E_{(s,a,r,s')\sim \mathcal{U}(\mathcal{D}) } \left[ \left(  y_i^{DQN} - Q(s,a; \theta_{i}) \right)^{2} \right].
\nonumber%\label{eq:lossreplay}
\ee
Experience replay increases data efficiency through re-use of experience samples in multiple updates and, importantly, it reduces variance as uniform sampling from the replay buffer reduces the correlation among the samples used in the update.

\subsection{Double Deep Q-networks} % <-- Love this section name. :) -- ML

The previous section described the main components of DQN as presented in \citep{Mnih:2015}. In this paper, we use the improved Double DQN (DDQN) learning algorithm of \citet{vanHasselt:2015}. In Q-learning and DQN, the $\max$ operator
uses the same values to both select and evaluate an action. This can therefore lead to overoptimistic value estimates \citep{vanHasselt:2010}. To mitigate this problem, DDQN uses the following target:
\be
y_i^{DDQN} =  r + \gamma Q(s',\argmax_{a'} Q(s',a';\theta_i)   ;\theta^{-}).
\ee
DDQN is the same as for DQN (see \citet{Mnih:2015}), but with the target $y_i^{DQN}$ replaced by $ y_i^{DDQN}$.
% Removed for ICML before accept.
The pseudo-code for DDQN is presented in Appendix~\ref{sec:ddqn_alg}.

\subsection{Prioritized Replay}

A recent innovation in prioritized experience replay~\citep{Schaul:2015} built on top of DDQN and further improved the state-of-the-art. Their key idea was to increase the replay probability of experience tuples that have a high expected learning progress (as measured via the proxy of absolute TD-error). This led to both faster learning and to better final policy quality across most games of the Atari benchmark suite, as compared to uniform experience replay.

To strengthen the claim that our dueling architecture is complementary to algorithmic innovations, we show that it improves performance for both the uniform and the prioritized replay baselines (for which we picked the easier to implement rank-based variant), with the resulting prioritized dueling variant holding the new state-of-the-art.

\section{The Dueling Network Architecture}
\label{sec:dueling}

%!TEX root = main.tex
%TODO(@Ziyu): Where are the nonlinearities? Also, did you use pooling? Did you explain more precisely in results section?

The key insight behind our new architecture, as illustrated in Figure~\ref{fig:saliency}, is that for many states, it is unnecessary to estimate the value of each action choice. For example, in the Enduro game setting, knowing whether to move left or right only matters when a collision is eminent. In some states, it is of paramount importance to know which action to take, but in many other states the choice of action has no repercussion on what happens. For bootstrapping based algorithms, however, the estimation of state values is of great importance for every state.

To bring this insight to fruition, we design a single $Q$-network architecture, as illustrated in Figure~\ref{fig:duelnet}, which we refer to as the dueling network. The lower layers of the dueling network are convolutional as in the original DQNs \cite{Mnih:2015}. However, instead of following the convolutional layers with a single sequence of fully connected layers, we instead use two sequences (or streams) of fully connected layers. The streams are constructed such that they have they have the capability of providing separate estimates of the value and advantage functions. Finally, the two streams are combined to produce a single output $Q$ function. As in \cite{Mnih:2015}, the output of the network is a set of $Q$ values, one for each action.

Since the output of the dueling network is a $Q$ function, it can be trained with the many existing algorithms, such as DDQN and SARSA. In addition, it can take advantage of any improvements to these algorithms, including better replay memories, better exploration policies, intrinsic motivation, and so on. 

The module that combines the two streams of fully-connected layers to output a $Q$ estimate requires very thoughtful design.  

From the expressions for advantage $Q^{\pi}(s,a) = V^{\pi}(s) + A^{\pi}(s,a)$ and state-value $V^{\pi}(s) = \expectQ{Q^{\pi}(s,a)}{a \sim \pi(s)}$, it follows that $\expectQ{A^{\pi}(s,a)}{a \sim \pi(s)} = 0$. Moreover, for a deterministic policy, $a^* = \argmax_{a' \in \mathcal{A}} Q(s,a')$, it follows that $Q(s,a^*) = V(s)$ and hence $A(s,a^*)=0$.

Let us consider the dueling network shown in Figure~\ref{fig:duelnet}, where we make one stream of fully-connected layers output a scalar ${V}(s;\theta,\beta)$, and the other stream output an $|\mathcal{A}|$-dimensional vector ${A}(s,a;\theta,\alpha)$. Here, $\theta$ denotes the parameters of the convolutional layers, while $\alpha$ and $\beta$ are the parameters of the two streams of fully-connected layers.

Using the definition of advantage, we might be tempted to construct the aggregating module as follows:
\be
Q(s,a;\theta,\alpha,\beta) =  {V}(s;\theta,\beta) + {A}(s,a;\theta,\alpha),
\label{eq:combo1}
\ee
Note that this expression applies to all $(s,a)$ instances; that is, to express equation~(\ref{eq:combo1}) in matrix form we need to replicate the scalar, ${V}(s;\theta,\beta)$, $|\mathcal{A}|$ times.

However, we need to keep in mind that $Q(s,a;\theta,\alpha,\beta)$ is only a parameterized estimate of the true $Q$-function. Moreover, it would be wrong to conclude
that ${V}(s;\theta,\beta)$ is a good estimator of the state-value function, or likewise that ${A}(s,a;\theta,\alpha)$ provides a reasonable estimate of the advantage function. 

Equation~(\ref{eq:combo1}) is unidentifiable in the sense that given $Q$ we cannot recover ${V}$ and ${A}$ uniquely. To see this, add a constant to ${V}(s;\theta,\beta)$ and subtract the same constant from ${A}(s,a;\theta,\alpha)$. This constant cancels out resulting in the same $Q$ value. 
% It is therefore not necessarily true that $V(s;\theta,\beta) = \max_a Q(s,a;\theta,\alpha,\beta)$ when acting according to the policy $Q(s,a;\theta,\alpha,\beta)$. 
This lack of identifiability is mirrored by poor practical performance when this equation is used directly.

To address this issue of identifiability, we can force the advantage function estimator to have zero advantage at the chosen action. That is, we let the last module of the network implement the forward mapping
\begin{multline}
Q(s,a;\theta,\alpha,\beta) =  {V}(s;\theta,\beta)~+\\
\left({A}(s,a;\theta,\alpha) - \max_{a' \in |\mathcal{A}|}  {A}(s, a' ;\theta,\alpha) \right).
\label{eq:combo3}
\end{multline}
Now, for $a^* = \argmax_{a' \in \mathcal{A}} Q(s,a';\theta,\alpha,\beta) = \argmax_{a' \in \mathcal{A}} A(s,a';\theta,\alpha)$, we obtain $Q(s,a^*;\theta,\alpha,\beta) =  {V}(s;\theta,\beta)$. Hence, the stream ${V}(s;\theta,\beta)$ provides an estimate of the value function, while the other stream produces an estimate of the advantage function.

An alternative module replaces the max operator with an average:
\begin{multline}
Q(s,a;\theta,\alpha,\beta) =  {V}(s;\theta,\beta)~+\\
\left({A}(s,a;\theta,\alpha) - \frac{1}{|\mathcal{A}|} \sum_{a'} {A}(s, a' ;\theta,\alpha) \right).
\label{eq:combo2}
\end{multline}
On the one hand this loses the original semantics of $V$ and $A$ because they are now off-target by a constant, but on the other hand it increases the stability of the optimization: with (\ref{eq:combo2}) the advantages only need to change as fast as the mean, instead of having to compensate any change to the optimal action's advantage in (\ref{eq:combo3}). We also experimented with a softmax version of equation (\ref{eq:combo3}), but found it to deliver similar results to the simpler module of equation (\ref{eq:combo2}). Hence, all the experiments reported in this paper use the module of equation (\ref{eq:combo2}). 

Note that while subtracting the mean in equation (\ref{eq:combo2}) helps with identifiability, it does not change the relative rank of the ${A}$ (and hence $Q$) values, preserving any greedy or $\epsilon$-greedy policy based on $Q$ values from equation (\ref{eq:combo1}).
When acting, it suffices to evaluate the advantage stream to make decisions.

It is important to note that equation (\ref{eq:combo2}) is viewed and implemented as part of the network and not as a separate algorithmic step. Training of the dueling architectures, as with standard $Q$ networks (e.g. the deep $Q$-network of \citet{Mnih:2015}), requires only back-propagation. The estimates ${V}(s;\theta,\beta)$ and ${A}(s,a;\theta,\alpha)$ are computed automatically without any extra supervision or algorithmic modifications. 

As the dueling architecture shares the same input-output interface with standard $Q$ networks, 
we can recycle all learning algorithms with $Q$ networks (\emph{e.g.}, DDQN and SARSA) to train the dueling architecture.

% With the dueling network, we learn the value stream with every update to the $Q$ values.
% This frequent updating, enables the dueling network to approximate the state
% values better.
% State value estimation is especially important to bootstrapping-based RL algorithms 
% \cite{SuttonBarto:1998}.

% The dueling architecture could also preserve the advantage orders
% while allowing changes in state value.

%\be
%Q(s,a;\theta,\alpha,\beta) =  V(s;\theta,\beta) + \left( A(s,a;\theta,\alpha) - \sum_{a'} \sigma[Q(s,a;\theta,%\alpha,\beta)] A(s,a';\theta,\alpha) \right) \label{eq:combo3} 
%\ee

%\be
%Q(s,a;\theta,\alpha,\beta) =  V(s;\theta,\beta) + \left( A(s,a;\theta,\alpha) - \frac{1}{N_a} \sum_{a'} A(s,a;\theta,\alpha) \right)
%\label{eq:combo4}
%\ee

%TODO: NNET EQUATION AND DIAGRAM

%TODO: GRADIENT CLIPPING, SOFTMAX, MAX, ETC.

\section{Experiments}
\label{sec:experiments}

%!TEX root = main.tex
%In this section, we evaluate our proposed approach on synthetic as 
%well as real world examples.

We now show the practical performance of the dueling network.
We start with a simple policy evaluation task and then show larger scale results
for learning policies for general Atari game-playing.

%\begin{wrapfigure}{R}{6cm}
% \begin{figure}[t]
% 	\centering
% 	\includegraphics[scale=0.3]{./figs/corridor}
% 	\caption{The corridor environment. The star marks the starting state. 
% 		The redness of a state signifies the reward the agent receives upon arrival. 
% 		The game terminates upon reaching either reward state. 
% 		The agent's actions are going up, down, left, right and no action.
% 		\label{fig:corridor}}
% \end{figure}
%\end{wrapfigure}

\subsection{Policy evaluation}

\begin{figure*}[t]
\begin{center}
	\begin{tabular}{cccc}
		{\sc Corridor environment}& {\sc ~~~~~~5 actions} & {\sc 10 actions} & {\sc 20 actions} \\
		\includegraphics[scale=0.22]{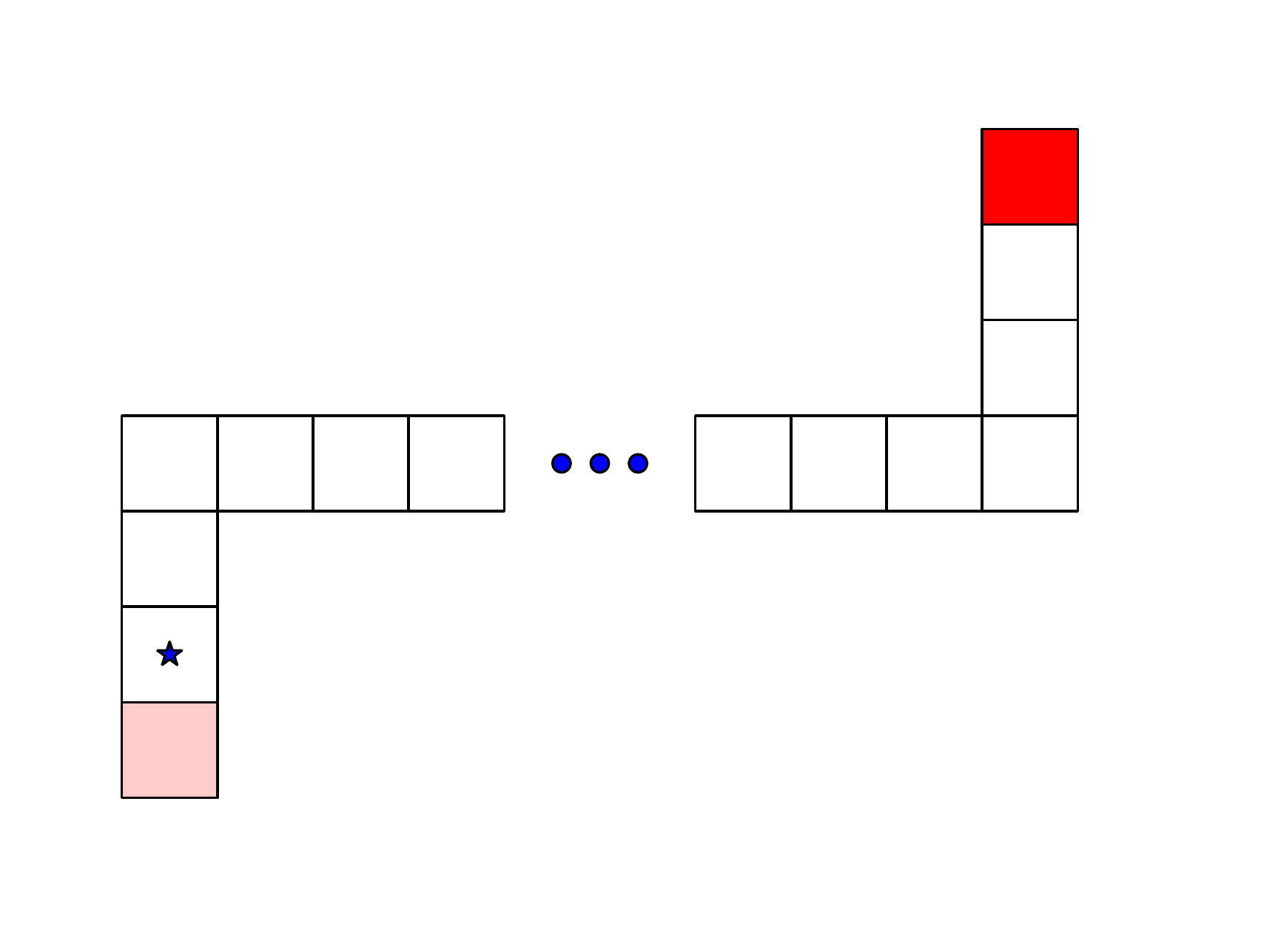} &
		\includegraphics[scale=0.25]{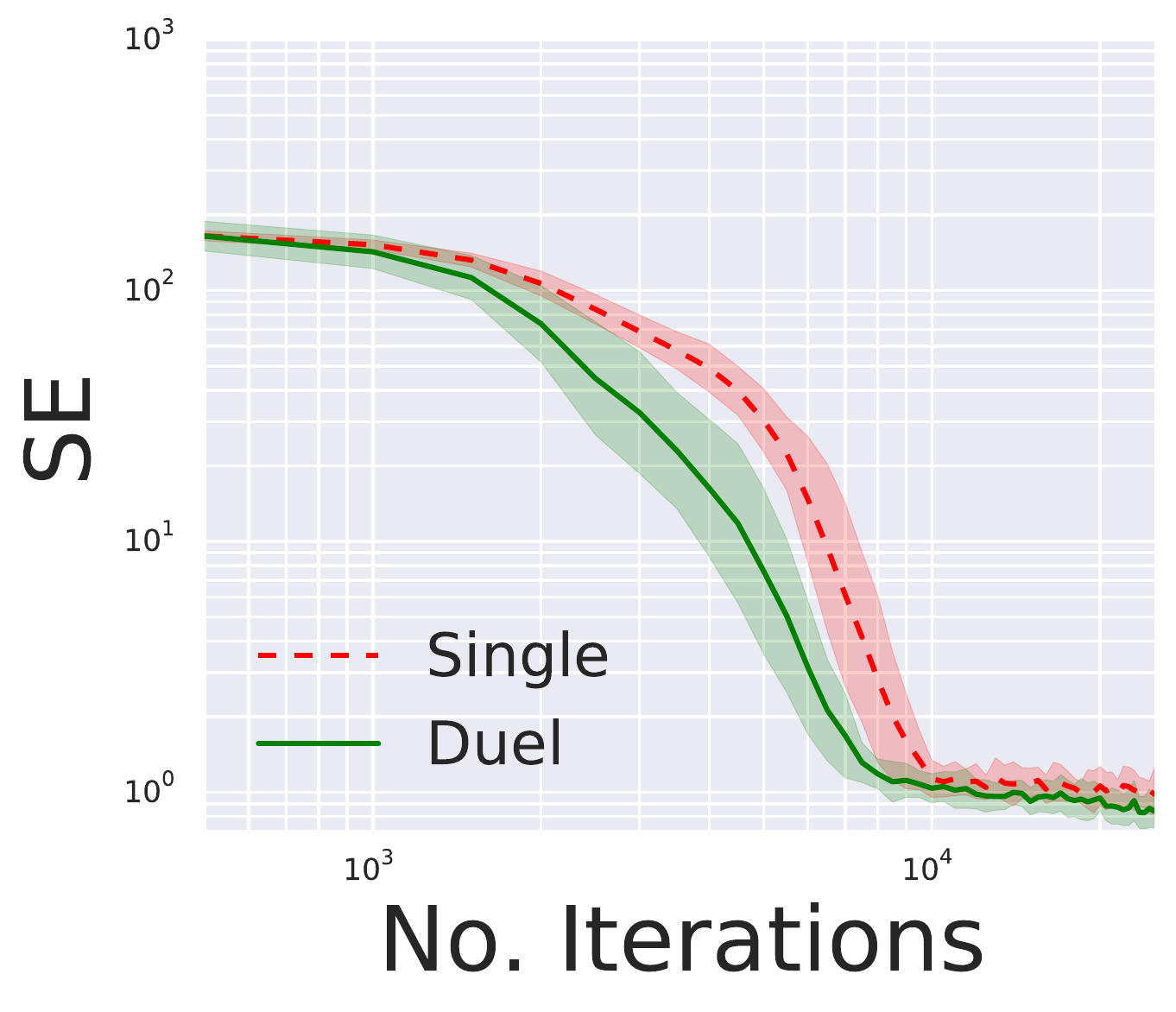} &
		\includegraphics[scale=0.25]{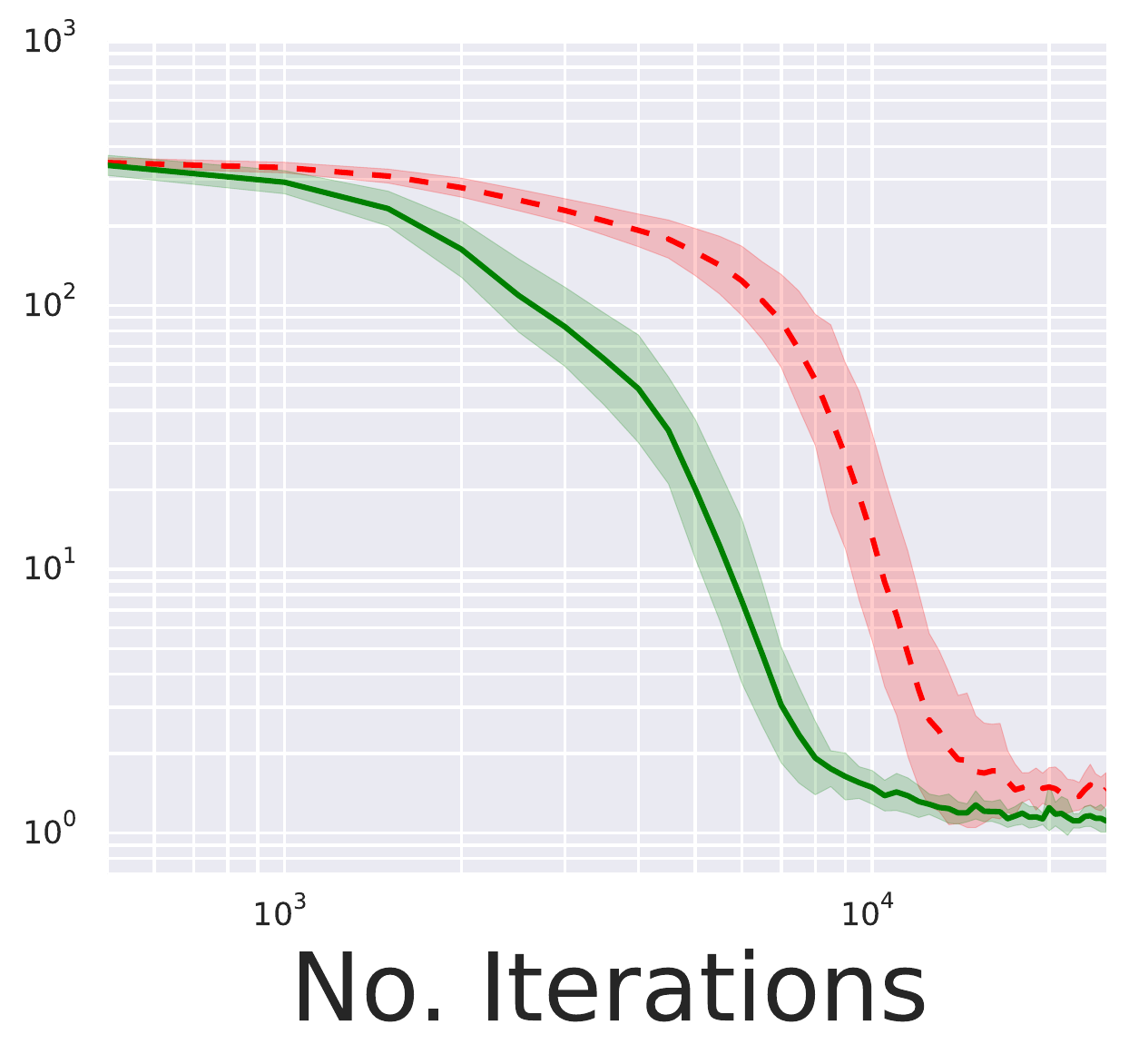}&
		\includegraphics[scale=0.25]{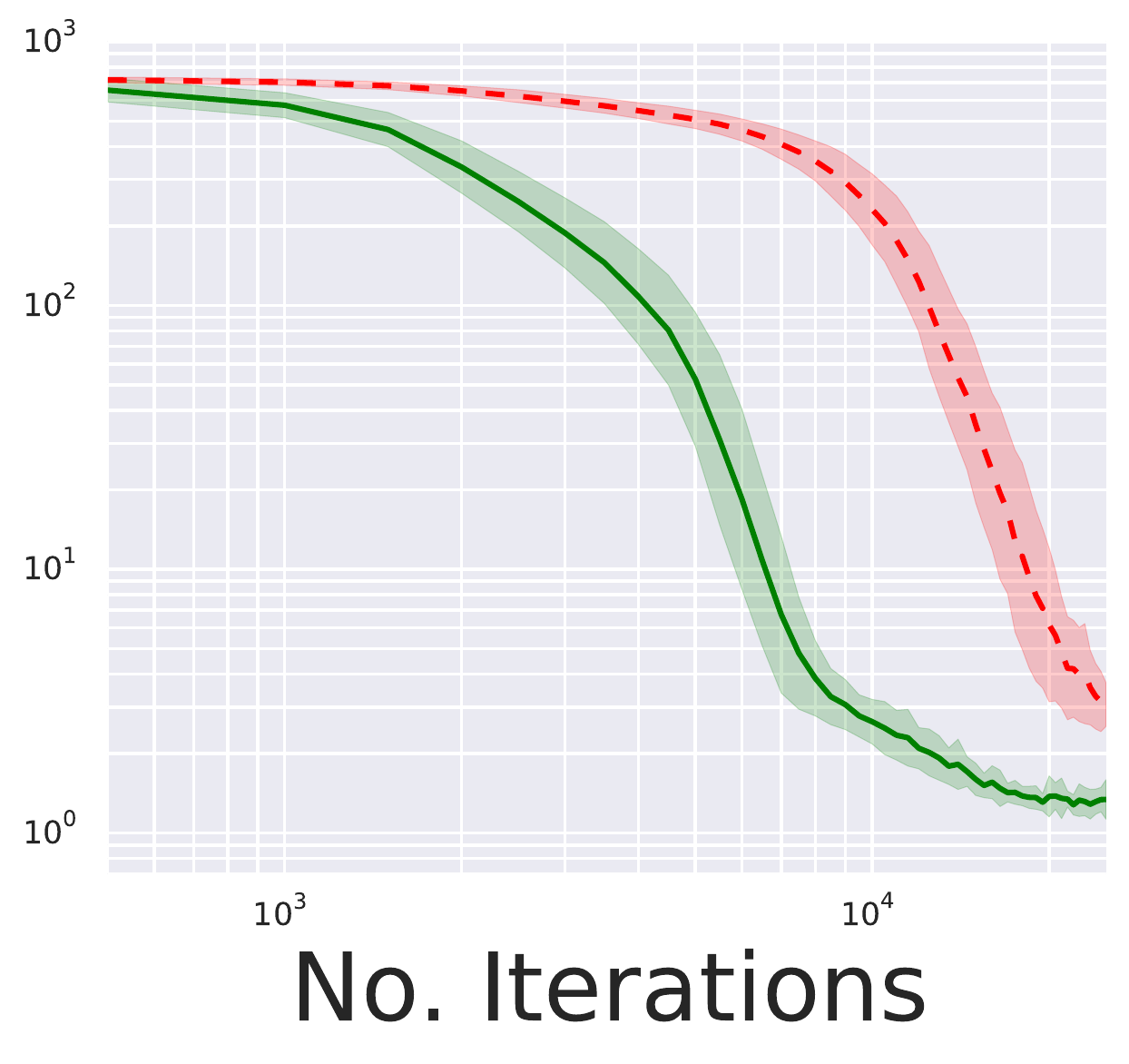} \\
		{\bf (a)}& {\bf (b)} & {\bf (c)} & {\bf (d)} 
	\end{tabular}
	\vspace{-1em}
\end{center}
\caption{{\bf (a)} The corridor environment. The star marks the starting state. 
		The redness of a state signifies the reward the agent receives upon arrival. 
		The game terminates upon reaching either reward state. 
		The agent's actions are going up, down, left, right and no action.
Plots {\bf(b), (c)} and {\bf(d)} shows squared error for policy evaluation with 5, 10, and 20 actions on a log-log scale. The dueling network (Duel) consistently outperforms a conventional single-stream network (Single), with the performance gap increasing with the number of actions.
	\label{fig:corrResults}}
\end{figure*}

We start by measuring the performance of the dueling architecture on a policy evaluation task.
%Compared to control tasks, we argue that 
We choose this particular task because it is very useful for evaluating network architectures,
as it is devoid of confounding factors 
such as the choice of exploration strategy, and the interaction between policy improvement and
policy evaluation.

In this experiment, we employ temporal difference learning (without eligibility traces, \emph{i.e.}, $\lambda = 0$) to learn $Q$ values.
More specifically, given a behavior policy $\pi$, we seek to estimate
the state-action value $Q^{\pi}(\cdot, \cdot)$ by optimizing the sequence of costs of equation~(\ref{eq:loss}), with target
%$$
%L_i(\theta_i) = \E_{s,a \sim \mathcal{B}} \left\{ \left( y_i - Q(s,a; \theta_{i}) \right)^{2} \right\},
%$$
%where $\mathcal{B}(s,a)$ is a behaviour distribution over states and actions, and the target is
\begin{equation*}
y_i =  r +  \gamma  \E_{a' \sim \pi(s')}\left[ Q(s',a';\theta_i) \right]. 
\end{equation*}
The above update rule is the same as that of Expected SARSA~\cite{vanSeijen:2009}.
We, however, do not modify the behavior policy as in Expected SARSA.

To evaluate the learned $Q$ values, 
we choose a simple environment 
where the exact $Q^\pi(s,a)$ values can be computed separately for all $(s,a) \in \cal{S} \times \cal{A}$.
This environment, which we call the {\it corridor} is composed of three connected corridors.
A schematic drawing of the corridor environment is shown in Figure~\ref{fig:corrResults}, 
The agent starts from the bottom left corner of the environment and must move to the top right to get the largest reward.
A total of $5$ actions are available: go up, down, left, right and
no-op. We also have the freedom of adding an arbitrary number of no-op actions.
In our setup, the two vertical sections both have 10 states while the horizontal 
section has 50.

We use an $\epsilon$-greedy policy as the behavior policy $\pi$, which chooses a random action with probability $\epsilon$ or an action according to the optimal $Q$ function
$\argmax_{a \in \cal{A}}Q^*(s,a)$ with probability $1-\epsilon$. In our experiments, $\epsilon$ is chosen to be $0.001$.

We compare a single-stream $Q$ architecture with the dueling architecture
on three variants of the corridor environment with 5, 10 and 20 actions 
respectively. The 10 and 20 action variants are formed by adding no-ops 
to the original environment.
We measure performance by Squared Error (SE) against the 
true state values: $\sum_{s \in \mathcal{S}, a \in \mathcal{A}} ( Q(s,a;\theta) - Q^{\pi}(s, a) )^2$.
The single-stream architecture is a three layer MLP with $50$ units
on each hidden layer. 
The dueling architecture is also composed of three layers.
After the first hidden layer of 50 units, however, the network branches
off into two streams each of them a two layer MLP with 25 hidden units.
The results of the comparison are summarized in Figure~\ref{fig:corrResults}.

The results show that with $5$ actions, both architectures converge at about the same speed.
However, when we increase the number of actions, the dueling
architecture performs better than the traditional $Q$-network. In the dueling network,
the stream ${V}(s;\theta,\beta)$ learns a
general value that is shared across many similar actions at $s$, hence leading to faster convergence. This is a very promising result because many control tasks with large action spaces have this property, and consequently we should expect that the dueling network will often lead to much faster convergence than a traditional single stream network. In the following section, we will indeed see that the dueling network results in substantial gains in performance in a wide-range of Atari games.

%which is a common
%occurrence in many control tasks.

\subsection{General Atari Game-Playing}

We perform a comprehensive evaluation of our proposed method on the Arcade Learning Environment~\cite{Bellemare:2013},
which is composed of 57 Atari games. The challenge is to deploy a single algorithm and architecture,
with a fixed set of hyper-parameters, to learn to play all the games
given only raw pixel observations and game rewards. 
This environment is very demanding because it is both comprised of a large number of highly diverse games and the observations are high-dimensional.

We follow closely the setup of~\citet{vanHasselt:2015} and compare
to their results using single-stream $Q$-networks. We train the dueling network with the DDQN algorithm as presented in Appendix~\ref{sec:ddqn_alg}. At the end of this section, we incorporate
prioritized experience replay~\cite{Schaul:2015}.

Our network architecture has the same low-level convolutional structure of DQN~\cite{Mnih:2015,vanHasselt:2015}.
There are 3 convolutional layers followed by 2 fully-connected layers. The first convolutional layer has 32 $8 \times 8$ filters with stride 4, the second 64 $4 \times 4$ filters with stride 2, and the third and final convolutional layer consists 64 $3 \times 3$ filters with stride 1. 
As shown in Figure~\ref{fig:duelnet},
the dueling network splits into two streams of fully connected layers. 
The value and advantage streams both have a fully-connected layer with 512 units. 
The final hidden layers of the value and advantage streams are both fully-connected
with the value stream having one output and the advantage as many outputs
as there are valid actions\footnote{The number of actions ranges between 3-18 actions in the ALE environment.}.
We combine the value and advantage streams using the module described by Equation~(\ref{eq:combo2}).
Rectifier non-linearities \cite{Fukushima:1980} are inserted between all adjacent layers.

%We refer to the popular convolutional Q-network
%architectures used in~\citet{Mnih:2015,vanHasselt:2015,Schaul:2015} as single stream networks.

We adopt the optimizers and hyper-parameters of~\citet{vanHasselt:2015},
with the exception of the learning rate which we chose to be slightly lower (we do not do this for double DQN as it can deteriorate its performance).
Since both the advantage and the value stream propagate gradients to the 
last convolutional layer in the backward pass, we rescale the combined gradient
entering the last convolutional layer by $1/\sqrt{2}$. This simple heuristic mildly increases stability.
In addition, we clip the gradients to have their norm less than or equal to 10. This clipping is not standard practice in deep RL, but common in recurrent network training \cite{Bengio:2013}. 

To isolate the contributions of the dueling architecture, we re-train DDQN
with a single stream network using exactly the same procedure 
as described above. Specifically, we apply gradient clipping, and
use $1024$ hidden units for the first fully-connected layer of the
network so that both architectures (dueling and single) have roughly the same
number of parameters. We refer to this re-trained model as \emph{Single Clip}, while the original trained model of~\citet{vanHasselt:2015} is referred to as \emph{Single}.

% During training, if the network produces outputs $\mathbf{o} = Q(s, \cdot;\theta,\alpha,\beta), \forall a \in \cal{A}$ for input $s$
% and we have associated targets $\mathbf{y}$ and
% $g = \nabla_{\theta,\alpha,\beta} Q(s,a;\theta,\alpha,\beta)$, then if $n = || g(s,\mathbf{o},\mathbf{y}) ||_2 \ge 10$ we instead use $g' = \frac{10g}{n}$.
% Gradient clipping, although not a commonly adopted practice
% for training convolutional networks, helps in the context of $Q$ learning.

% All three architectures are trained with the same hyper-parameters.
% As shown in Figure~\ref{fig:trainCurves}, 
% gradient clipping clearly helps with training. 
% The dueling architecture still maintains
% a very significant advantage over the traditional $Q$ architecture 
% in the presence of gradient clipping.
% Overall, the dueling architecture does equally or better on 75.4\% of the games (43 out of 57). 

\begin{figure}[t!]
\begin{center}
	\includegraphics[scale=0.47]{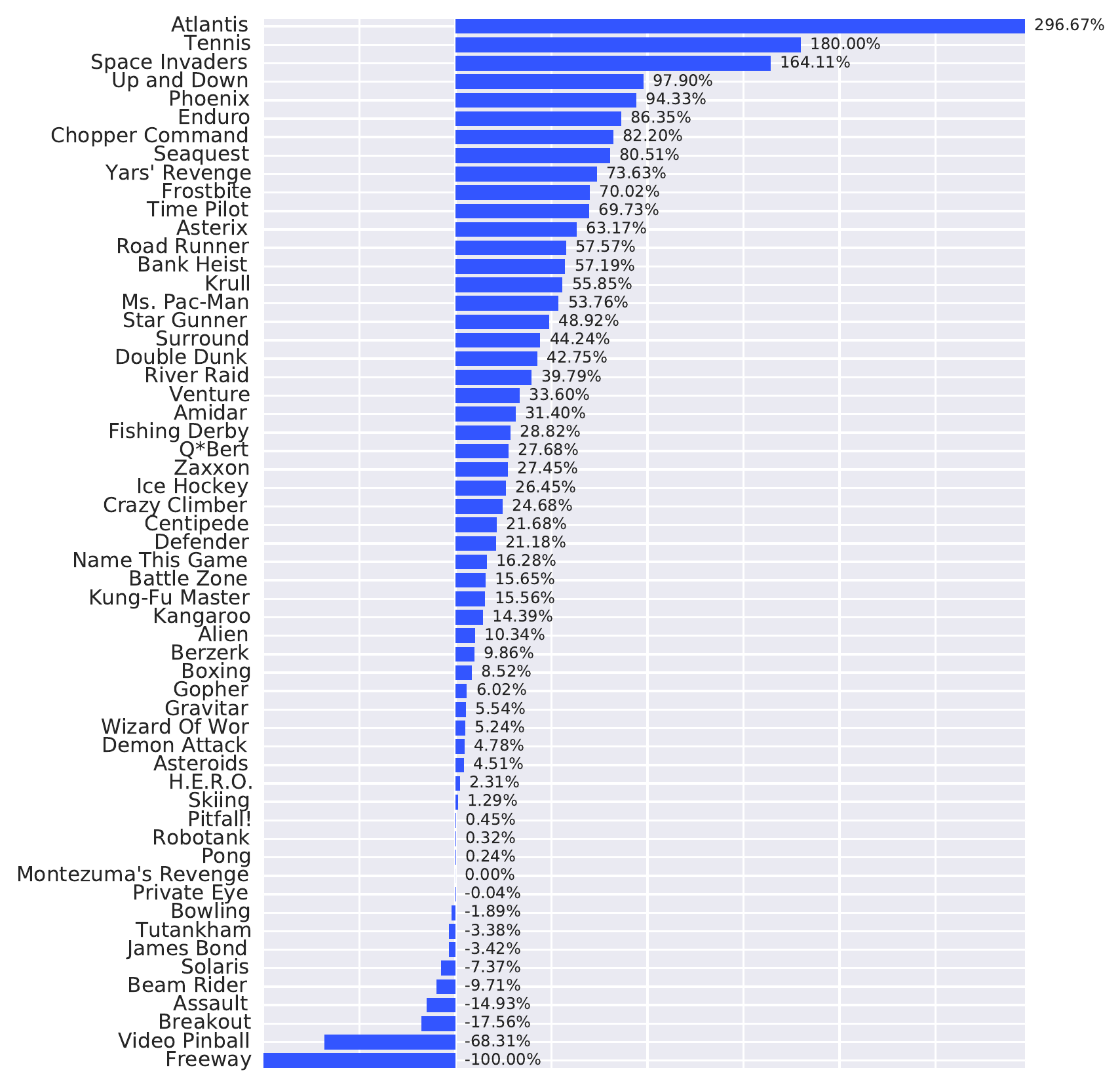}
\end{center}
\caption{\label{fig:at30} Improvements of dueling architecture over the baseline Single network of~\citet{vanHasselt:2015}, using the metric described in Equation~(\ref{eq:metric}). Bars to the right indicate by how much the dueling network outperforms the single-stream network.}
\end{figure}
%\begin{wrapfigure}{R}{5cm}
%\vspace{-30pt}

%\end{wrapfigure}

As in~\cite{vanHasselt:2015}, 
we start the game with up to \emph{30 no-op} actions to provide random starting positions for the agent.
To evaluate our approach, we measure improvement in percentage (positive or negative) 
in score over the better of human and baseline agent scores: 
\be
\label{eq:metric}
{\footnotesize
\frac{\mbox{\small Score}_{\mbox{\tiny Agent}} - \mbox{\small Score}_{\mbox{\tiny Baseline}}}
	{\small
		\max\{
			\mbox{Score}_{\mbox{\tiny Human}}, \mbox{Score}_{\mbox{\tiny Baseline}} 
		 \}
	-  \mbox{Score}_{\mbox{\tiny Random}}
	}.
}
\ee
We took the maximum over human and baseline agent scores as it prevents insignificant 
changes to appear as large improvements 
when neither the agent in question nor the baseline are doing well.
For example, an agent that achieves 2\% human performance should not be interpreted 
as two times better when the baseline agent achieves 1\% human performance.
We also chose not to measure performance in terms of percentage of human performance alone
because a tiny difference relative to the baseline on some games can translate into hundreds of
percent in human performance difference.

The results for the wide suite of 57 games are summarized in~Table~\ref{tb:result}. Detailed results are presented in the Appendix.

Using this 30 no-ops performance measure, it is clear that the dueling network (Duel Clip) does substantially better than the Single Clip network of similar capacity. It also does considerably better than the baseline (Single) of~\citet{vanHasselt:2015}. For comparison we also show results for the deep $Q$-network of \citet{Mnih:2015}, referred to as Nature DQN.

Figure~\ref{fig:at30} shows the improvement of the dueling network over the baseline Single network of~\citet{vanHasselt:2015}. Again, we seen that the improvements are often very dramatic.

As shown in Table~\ref{tb:result}, Single Clip performs better than Single. We verified that this gain was mostly brought in by gradient clipping. For this reason, we incorporate gradient clipping in all the new approaches.

%\footnote{	When calculating the human performance percentage for the game	Video Pinball, the random scores are set to zero. 	This is because human scores 	on this game are lower than that of a random agent making the calculation of	human performance percentage invalid.	Resetting the random score does not affect the median scores of the agents	considered. The mean scores, however, are affected.}

\begin{table}[t]
\caption{Mean and median scores across all 57 Atari games, measured in percentages of human performance.}
\label{tb:result}
\small
\begin{center}
\begin{tabular}{l|rr|rr}
\multicolumn{1}{c}{\bf }  &\multicolumn{2}{c}{\bf 30 no-ops} &\multicolumn{2}{c}{\bf Human Starts} \\
 \hline
\multicolumn{1}{c}{\bf }  &\multicolumn{1}{r}{\bf Mean} &\multicolumn{1}{r}{\bf Median}&\multicolumn{1}{r}{\bf Mean} &\multicolumn{1}{r}{\bf Median}
\\ \hline 
Prior. Duel Clip  &  {\bf 591.9}\%  & {\bf 172.1}\% &  {\bf 567.0}\%  & {\bf 115.3}\% \\
Prior. Single&  434.6\%  & 123.7\% &  386.7\%  & 112.9\% \\
\hline
Duel Clip       &  {\bf 373.1}\%  & {\bf 151.5}\% &  {\bf 343.8}\%  & {\bf 117.1}\% \\
Single Clip  &  341.2\%  & 132.6\% &  302.8\%  & 114.1\% \\
Single            &  307.3\%   & 117.8\% &  332.9\%   & 110.9\% \\
\hline
Nature DQN            &  227.9\%   & 79.1\% &  219.6\%   & 68.5\% 
\end{tabular}
\end{center}
\end{table}

Duel Clip does better than Single Clip 
on 75.4\% of the games (43 out of 57). It also achieves higher scores compared to the Single baseline on 80.7\% ($46$
out of $57$) of the games. Of all the games with $18$ actions, 
Duel Clip is better $86.6$\% of the time (26 out of 30).
This is consistent with the findings of the previous section.
Overall, our agent (Duel Clip) achieves human level performance on 42 out of 57 games.
Raw scores for all the games, as well as measurements in human performance percentage,
are presented in the Appendix.
% In Figure~\ref{fig:natureComp}, we show a comparison between our
% agent, DQN~\cite{mnih2015human}, and human expert.

% A few training curves, shown in Figure~\ref{fig:trainCurves}, 
% compare the dueling architecture, and DDQN with the traditional $Q$ architecture with
% and without gradient clipping.
% \begin{figure}[h]
% \begin{center}
% 	\includegraphics[scale=0.19]{./figs/enduro}
% 	\includegraphics[scale=0.19]{./figs/kung_fu_master}
% 	\includegraphics[scale=0.19]{./figs/riverraid} 
% 	\includegraphics[scale=0.19]{./figs/seaquest}
% \end{center}
% \caption{Performance evaluated at training time on four Atari games. The curves correspond to three instances of the DDQN algorithm using a convolutional network of similar size to the dueling network (Q net) using the same convolutional network and gradient clipping (Q net Clip) and using the dueling network (Duel).  
% 	\label{fig:trainCurves}}
% \end{figure}
% Seaquest $0.039$ $0.039$
% Ms. Pacman $0.032$ $0.042$

% \subsubsection*{Human Random Starts}
{\bf Robustness to human starts.}
One shortcoming of the 30 no-ops metric is that an agent does not necessarily have
to generalize well to play the Atari games.
Due to the deterministic nature of the Atari environment, 
from an unique starting point, an agent could learn to achieve good performance by simply 
remembering sequences of actions.

To obtain a more robust measure, we adopt the methodology of 
~\citet{Nair:2015}.
Specifically, for each game, we use 100 starting points sampled from a human expert's trajectory.
From each of these points, an evaluation episode is launched for up to 108,000 frames.
The agents are evaluated only on rewards accrued after the starting point. We refer to this metric as \emph{Human Starts}.

As shown in Table~\ref{tb:result}, under the Human Starts metric, Duel Clip once again outperforms the single stream variants. In particular, our agent does better than the Single baseline on 70.2\% (40 out of 57) games
and on games of $18$ actions, Duel Clip is 83.3\% better (25 out of 30).

\begin{figure}[t!]
\begin{center}
	\includegraphics[scale=0.47]{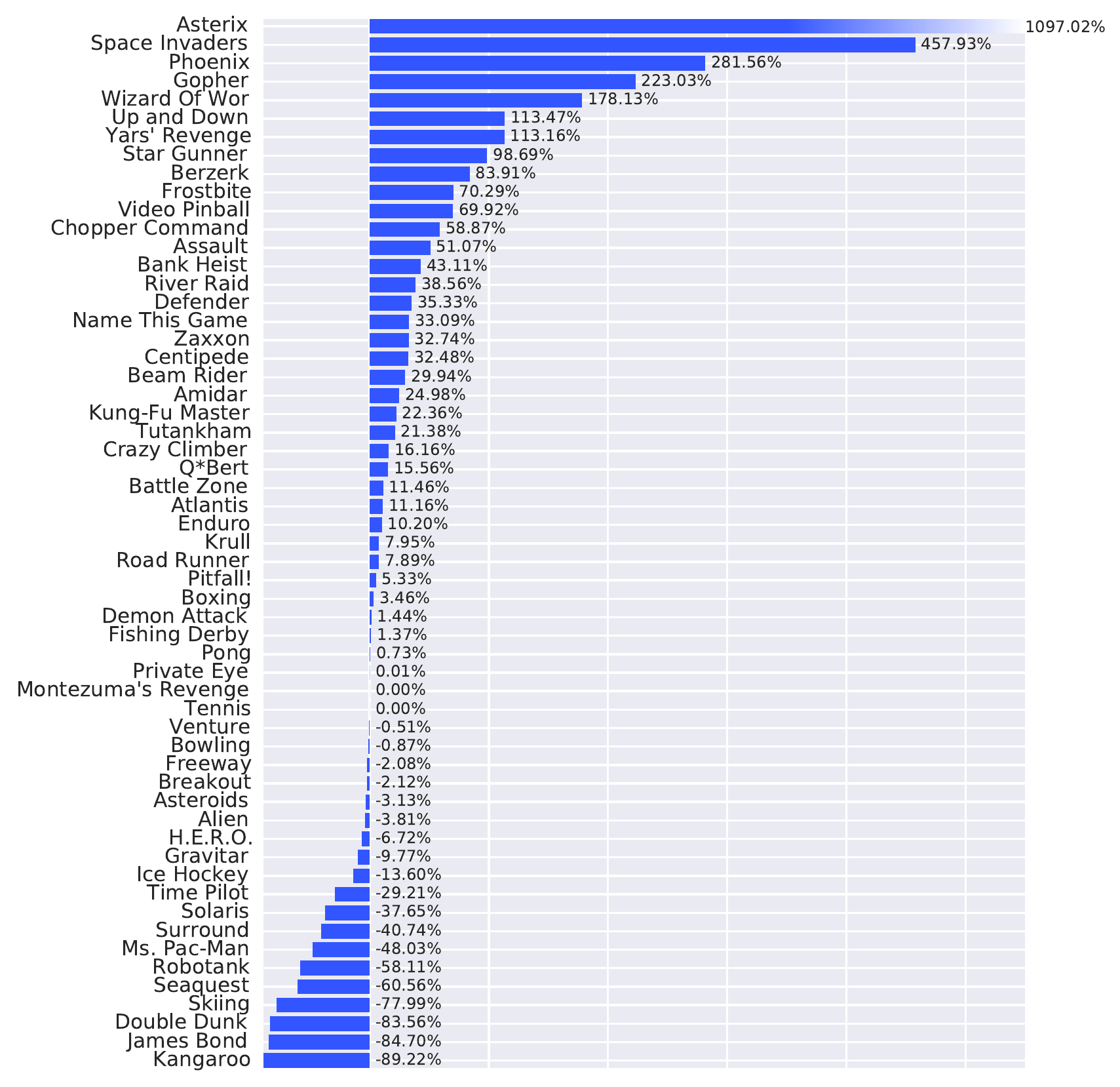}
\end{center}
\caption{\label{fig:at30_priduel} 
Improvements of dueling architecture over Prioritized DDQN baseline, using the same metric as Figure~\ref{fig:at30}. 
Again, the dueling architecture leads to significant improvements over the single-stream baseline on the majority of games.
}
\end{figure}

{\bf Combining with Prioritized Experience Replay.}
\label{sec:prioritizedduel}
The dueling architecture can be easily combined with other algorithmic improvements. 
In particular, prioritization of the experience replay has been shown to significantly improve performance of Atari games~\cite{Schaul:2015}. 
Furthermore, as prioritization and the dueling architecture address very different aspects of the learning process, their combination is promising. 
So in our final experiment, we investigate the integration of the dueling architecture with prioritized experience replay.
We use the prioritized variant of DDQN (Prior. Single) as the new baseline algorithm, which replaces with the uniform sampling
of the experience tuples by rank-based prioritized sampling. 
We keep all the parameters of the prioritized replay as described in~\cite{Schaul:2015}, namely
a priority exponent of $0.7$, and an annealing schedule on the importance sampling exponent from $0.5$ to $1$.
% , as well as sorting the heap every $10^6$ frames. %TODO
We combine this baseline with our dueling architecture (as above), and again use gradient clipping (Prior. Duel Clip).

Note that, although orthogonal in their objectives, these extensions (prioritization, dueling and gradient clipping) interact in subtle ways. 
For example, prioritization interacts with gradient clipping, as sampling transitions with high absolute TD-errors more often leads to gradients with higher norms. 
To avoid adverse interactions, we roughly re-tuned the learning rate and the gradient clipping norm on a subset of $9$ games. As a result of rough tuning, we settled on $6.25 \times 10^{-5}$ for the learning rate and $10$ for the gradient clipping norm (the same as in the previous section).

When evaluated on all $57$ Atari games, our prioritized dueling agent performs significantly better than both the prioritized baseline agent and the dueling agent alone.
The full mean and median performance against the human performance percentage is shown in Table~\ref{tb:result}. When initializing the games using up to 30 no-ops action, we observe mean and median scores of 591\% and 172\% respectively.
The direct comparison between the prioritized baseline and prioritized dueling versions, using the metric described in Equation~\ref{eq:metric}, is presented in Figure~\ref{fig:at30_priduel}.

The combination of prioritized replay and the dueling network results in vast improvements over the previous state-of-the-art in the popular ALE benchmark.

% When initializing the games using up to 30 no-ops, we observed 591\% mean, and 172\% median, normalized scores. This agent also achieves the best reported performance on 6 Atari games (of which 5 were not in the set of games used for tuning). 

% subsection  (end)

{\bf Saliency maps.} To better understand the roles of the value
and the advantage streams, we compute saliency maps~\cite{Simonyan:2013}.
More specifically, to visualize the salient part of the image as seen by the value stream,
we compute the absolute value of the Jacobian of $\widehat{V}$ with respect to the input frames:
$|\nabla_{s} \widehat{V}(s ;\theta)|.$
Similarly, to visualize the salient part of the image as seen by the advantage stream,
we compute $|\nabla_{s} \widehat{A}(s, \argmax_{a'} \widehat{A}(s, a');\theta)|$.
Both quantities are of the same dimensionality as the input frames and therefore
can be visualized easily alongside the input frames.

Here, we place the gray scale input frames in the green and blue channel and
the saliency maps in the red channel.
All three channels together form an RGB image.
Figure~\ref{fig:saliency} depicts the
value and advantage saliency maps on the Enduro game for two different time steps. 
As observed in the introduction, the value stream pays attention 
to the horizon where the appearance of a car could affect future performance.
The value stream also pays attention to the score.
The advantage stream, on the other hand, cares more about cars 
that are on an immediate collision course.

\section{Discussion}
\label{sec:discussion}
The advantage of the dueling architecture lies partly in its ability to learn the state-value function efficiently.
With every update of the $Q$ values in the dueling architecture,
the value stream ${V}$ is updated -- this contrasts with the updates in a single-stream architecture where only the value for one of the actions is updated, the values for all other actions remain untouched. This more frequent updating of the value stream in our approach allocates more resources to $V$, and thus allows for better approximation of the state values, which in turn need to be accurate for temporal-difference-based methods like Q-learning to work~\cite{SuttonBarto:1998}.
This phenomenon is reflected in the experiments, where the advantage of the dueling architecture over single-stream 	$Q$ networks grows when the number of actions is large.

Furthermore, the differences between $Q$-values for a given state are often very small relative to the magnitude of $Q$.
For example, after training with DDQN on the game of Seaquest, the average action gap (the gap between the $Q$ values of the best and the second best action in a given state) across visited states is roughly $0.04$, whereas the average state value across those states is about $15$.
This difference in scales can lead to small amounts of noise in the updates can lead to reorderings of the actions, and thus make the nearly greedy policy switch abruptly.
The dueling architecture with its separate advantage stream is robust to such effects.

% Max vs. Mean

\section{Conclusions}
\label{sec:conclusion}

We introduced a new neural network architecture that decouples value and advantage in deep $Q$-networks, while sharing a common feature learning module. The new dueling architecture, in combination with some algorithmic improvements, leads to dramatic improvements over existing approaches for deep RL in the challenging Atari domain. The results presented in this paper are the new state-of-the-art in this popular domain. 

% Add after accepted
%\subsubsection*{Acknowledgments}
%We would like to thank Hado van Hasselt, Arthur Guez, Vlad Mnih, Nicolas Hess, Marc Bellemare, Georg Ostrovski, Tom Schaul and all the folks at Google DeepMind for making this possible.

\bibliography{deeprl}
\bibliographystyle{icml2016}

\appendix

\onecolumn
\section{Double DQN Algorithm}
\label{sec:ddqn_alg}
%!TEX root = main.tex
\begin{algorithm2e}[h!]
\small
\SetKwInOut{Input}{input}\SetKwInOut{Output}{output}
\Input{$\mathcal{D}$ -- empty replay buffer; $\theta$ -- initial network parameters, $\theta^-$ -- copy of $\theta$}
\Input{$N_r$ -- replay buffer maximum size; $N_b$ -- training batch size; $N^-$ -- target network replacement freq.}
\For{\mbox{episode} $e \in \{ 1, 2, \ldots, M$ \} } {
  Initialize frame sequence $\mathbf{x} \leftarrow ()$ \;
  \For{$t \in \{ 0, 1, \ldots \} $} {
    Set state $s \leftarrow \mathbf{x}$, sample action $a \sim \pi_\mathcal{B}$ \;
    Sample next frame $x^t$ from environment $\mathcal{E}$ given $(s,a)$ and receive reward $r$, and append $x^t$ to $\mathbf{x}$ \;
    {\bf if} {$|\mathbf{x}| > N_f$} {\bf then} delete oldest frame $x_{t_{min}}$ from $\mathbf{x}$ {\bf end} \;
    Set $s' \leftarrow \mathbf{x}$, and add transition tuple $(s, a, r, s')$ to $\mathcal{D}$,\\~~~~~~~~~replacing the oldest tuple if $|\mathcal{D}| \ge N_r$ \;
    Sample a minibatch of $N_b$ tuples $(s, a, r, s') \sim \mbox{Unif}(\mathcal{D})$\; %uniformly from $\mathcal{D}$ \;
    Construct target values, one for each of the $N_b$ tuples: \;
    Define $a^{\max{}}(s'; \theta) = \argmax_{a'} Q(s', a'; \theta)$\;
    $y_j = \left\{ \begin{array}{ll}
         r & \mbox{if $s'$ is terminal}\\
         r + \gamma Q(s', a^{\max{}}(s'; \theta); \theta^-), & \mbox{otherwise}. \end{array} \right.$ \;
    Do a gradient descent step with loss $\|y_j - Q(s, a ; \theta )\|^2$ \;
    Replace target parameters $\theta^- \leftarrow \theta$ every $N^-$ steps\;
  }
}
\caption{Double DQN Algorithm.}\label{alg:ddqn}
\end{algorithm2e}

%!TEX root = main.tex
\label{sec:appen}
\begin{table}[!b]
\footnotesize
\caption{Raw scores across all games. Starting with {\bf 30 no-op} actions.}
\begin{center}
\begin{tabular}{l|rrr|rrr|rr}
{\sc Games} &  {\sc No. Actions} &     {\sc Random} &     {\sc Human} &        {\sc DQN} &       {\sc DDQN} &       {\sc Duel} &{\sc Prior.} & {\sc Prior. Duel.}\\
\hline
               Alien &              18 &      227.8 &   7,127.7 &    1,620.0 &    3,747.7 &{\bf4,461.4}&    4,203.8 &      3,941.0 \\
              Amidar &              10 &        5.8 &   1,719.5 &      978.0 &    1,793.3 &{\bf2,354.5}&    1,838.9 &      2,296.8 \\
             Assault &               7 &      222.4 &     742.0 &    4,280.4 &    5,393.2 &    4,621.0 &    7,672.1 &{\bf11,477.0}\\
             Asterix &               9 &      210.0 &   8,503.3 &    4,359.0 &   17,356.5 &   28,188.0 &   31,527.0 &{\bf375,080.0}\\
           Asteroids &              14 &      719.1 &  47,388.7 &    1,364.5 &      734.7 &{\bf2,837.7}&    2,654.3 &      1,192.7 \\
            Atlantis &               4 &   12,850.0 &  29,028.1 &  279,987.0 &  106,056.0 &  382,572.0 &  357,324.0 &{\bf395,762.0}\\
          Bank Heist &              18 &       14.2 &     753.1 &      455.0 &    1,030.6 &{\bf1,611.9}&    1,054.6 &      1,503.1 \\
         Battle Zone &              18 &    2,360.0 &  37,187.5 &   29,900.0 &   31,700.0 &{\bf37,150.0}&   31,530.0 &     35,520.0 \\
          Beam Rider &               9 &      363.9 &  16,926.5 &    8,627.5 &   13,772.8 &   12,164.0 &   23,384.2 &{\bf30,276.5}\\
             Berzerk &              18 &      123.7 &   2,630.4 &      585.6 &    1,225.4 &    1,472.6 &    1,305.6 &{\bf3,409.0}\\
             Bowling &               6 &       23.1 &     160.7 &       50.4 &{\bf68.1}&       65.5 &       47.9 &         46.7 \\
              Boxing &              18 &        0.1 &      12.1 &       88.0 &       91.6 &{\bf99.4}&       95.6 &         98.9 \\
            Breakout &               4 &        1.7 &      30.5 &      385.5 &{\bf418.5}&      345.3 &      373.9 &        366.0 \\
           Centipede &              18 &    2,090.9 &  12,017.0 &    4,657.7 &    5,409.4 &    7,561.4 &    4,463.2 &{\bf7,687.5}\\
     Chopper Command &              18 &      811.0 &   7,387.8 &    6,126.0 &    5,809.0 &   11,215.0 &    8,600.0 &{\bf13,185.0}\\
       Crazy Climber &               9 &   10,780.5 &  35,829.4 &  110,763.0 &  117,282.0 &  143,570.0 &  141,161.0 &{\bf162,224.0}\\
            Defender &              18 &    2,874.5 &  18,688.9 &   23,633.0 &   35,338.5 &{\bf42,214.0}&   31,286.5 &     41,324.5 \\
        Demon Attack &               6 &      152.1 &   1,971.0 &   12,149.4 &   58,044.2 &   60,813.3 &   71,846.4 &{\bf72,878.6}\\
         Double Dunk &              18 &      -18.6 &     -16.4 &       -6.6 &       -5.5 &        0.1 &{\bf18.5}&        -12.5 \\
              Enduro &               9 &        0.0 &     860.5 &      729.0 &    1,211.8 &    2,258.2 &    2,093.0 &{\bf2,306.4}\\
       Fishing Derby &              18 &      -91.7 &     -38.7 &       -4.9 &       15.5 &{\bf46.4}&       39.5 &         41.3 \\
             Freeway &               3 &        0.0 &      29.6 &       30.8 &       33.3 &        0.0 &{\bf33.7}&         33.0 \\
           Frostbite &              18 &       65.2 &   4,334.7 &      797.4 &    1,683.3 &    4,672.8 &    4,380.1 &{\bf7,413.0}\\
              Gopher &               8 &      257.6 &   2,412.5 &    8,777.4 &   14,840.8 &   15,718.4 &   32,487.2 &{\bf104,368.2}\\
            Gravitar &              18 &      173.0 &   3,351.4 &      473.0 &      412.0 &{\bf588.0}&      548.5 &        238.0 \\
            H.E.R.O. &              18 &    1,027.0 &  30,826.4 &   20,437.8 &   20,130.2 &   20,818.2 &{\bf23,037.7}&     21,036.5 \\
          Ice Hockey &              18 &      -11.2 &       0.9 &       -1.9 &       -2.7 &        0.5 &{\bf1.3}&         -0.4 \\
          James Bond &              18 &       29.0 &     302.8 &      768.5 &    1,358.0 &    1,312.5 &{\bf5,148.0}&        812.0 \\
            Kangaroo &              18 &       52.0 &   3,035.0 &    7,259.0 &   12,992.0 &   14,854.0 &{\bf16,200.0}&      1,792.0 \\
               Krull &              18 &    1,598.0 &   2,665.5 &    8,422.3 &    7,920.5 &{\bf11,451.9}&    9,728.0 &     10,374.4 \\
      Kung-Fu Master &              14 &      258.5 &  22,736.3 &   26,059.0 &   29,710.0 &   34,294.0 &   39,581.0 &{\bf48,375.0}\\
 Montezuma's Revenge &              18 &        0.0 &   4,753.3 &{\bf0.0}&        0.0 &        0.0 &        0.0 &          0.0 \\
         Ms. Pac-Man &               9 &      307.3 &   6,951.6 &    3,085.6 &    2,711.4 &    6,283.5 &{\bf6,518.7}&      3,327.3 \\
      Name This Game &               6 &    2,292.3 &   8,049.0 &    8,207.8 &   10,616.0 &   11,971.1 &   12,270.5 &{\bf15,572.5}\\
             Phoenix &               8 &      761.4 &   7,242.6 &    8,485.2 &   12,252.5 &   23,092.2 &   18,992.7 &{\bf70,324.3}\\
            Pitfall! &              18 &     -229.4 &   6,463.7 &     -286.1 &      -29.9 &{\bf0.0}&     -356.5 &          0.0 \\
                Pong &               3 &      -20.7 &      14.6 &       19.5 &       20.9 &{\bf21.0}&       20.6 &         20.9 \\
         Private Eye &              18 &       24.9 &  69,571.3 &      146.7 &      129.7 &      103.0 &      200.0 &{\bf206.0}\\
              Q*Bert &               6 &      163.9 &  13,455.0 &   13,117.3 &   15,088.5 &{\bf19,220.3}&   16,256.5 &     18,760.3 \\
          River Raid &              18 &    1,338.5 &  17,118.0 &    7,377.6 &   14,884.5 &{\bf21,162.6}&   14,522.3 &     20,607.6 \\
         Road Runner &              18 &       11.5 &   7,845.0 &   39,544.0 &   44,127.0 &{\bf69,524.0}&   57,608.0 &     62,151.0 \\
            Robotank &              18 &        2.2 &      11.9 &       63.9 &       65.1 &{\bf65.3}&       62.6 &         27.5 \\
            Seaquest &              18 &       68.4 &  42,054.7 &    5,860.6 &   16,452.7 &{\bf50,254.2}&   26,357.8 &        931.6 \\
              Skiing &               3 &  -17,098.1 &  -4,336.9 &  -13,062.3 &   -9,021.8 &{\bf-8,857.4}&   -9,996.9 &    -19,949.9 \\
             Solaris &              18 &    1,236.3 &  12,326.7 &    3,482.8 &    3,067.8 &    2,250.8 &{\bf4,309.0}&        133.4 \\
      Space Invaders &               6 &      148.0 &   1,668.7 &    1,692.3 &    2,525.5 &    6,427.3 &    2,865.8 &{\bf15,311.5}\\
         Star Gunner &              18 &      664.0 &  10,250.0 &   54,282.0 &   60,142.0 &   89,238.0 &   63,302.0 &{\bf125,117.0}\\
           Surround  &               5 &      -10.0 &       6.5 &       -5.6 &       -2.9 &        4.4 &{\bf8.9}&          1.2 \\
              Tennis &              18 &      -23.8 &      -8.3 &{\bf12.2}&      -22.8 &        5.1 &        0.0 &          0.0 \\
          Time Pilot &              10 &    3,568.0 &   5,229.2 &    4,870.0 &    8,339.0 &{\bf11,666.0}&    9,197.0 &      7,553.0 \\
           Tutankham &               8 &       11.4 &     167.6 &       68.1 &      218.4 &      211.4 &      204.6 &{\bf245.9}\\
         Up and Down &               6 &      533.4 &  11,693.2 &    9,989.9 &   22,972.2 &{\bf44,939.6}&   16,154.1 &     33,879.1 \\
             Venture &              18 &        0.0 &   1,187.5 &      163.0 &       98.0 &{\bf497.0}&       54.0 &         48.0 \\
       Video Pinball &               9 &   16,256.9 &  17,667.9 &  196,760.4 &  309,941.9 &   98,209.5 &  282,007.3 &{\bf479,197.0}\\
       Wizard Of Wor &              10 &      563.5 &   4,756.5 &    2,704.0 &    7,492.0 &    7,855.0 &    4,802.0 &{\bf12,352.0}\\
       Yars' Revenge &              18 &    3,092.9 &  54,576.9 &   18,098.9 &   11,712.6 &   49,622.1 &   11,357.0 &{\bf69,618.1}\\
              Zaxxon &              18 &       32.5 &   9,173.3 &    5,363.0 &   10,163.0 &   12,944.0 &   10,469.0 &{\bf13,886.0}\\
\end{tabular}
\end{center}
\end{table}

\begin{table}[t]
\caption{Raw scores across all games. Starting with {\bf Human starts}.}
\footnotesize
\begin{center}
\begin{tabular}{l|rrr|rrr|rr}
    {\sc Games} &  {\sc No. Actions} &     {\sc Random} &     {\sc Human} &        {\sc DQN} &       {\sc DDQN} &       {\sc Duel} &{\sc Prior.} & {\sc Prior. Duel.}\\
\hline
               Alien &              18 &      128.3 &   6,371.3 &      634.0 &    1,033.4 &{\bf1,486.5}&    1,334.7 &        823.7 \\
              Amidar &              10 &       11.8 &   1,540.4 &      178.4 &      169.1 &      172.7 &      129.1 &{\bf238.4}\\
             Assault &               7 &      166.9 &     628.9 &    3,489.3 &    6,060.8 &    3,994.8 &    6,548.9 &{\bf10,950.6}\\
             Asterix &               9 &      164.5 &   7,536.0 &    3,170.5 &   16,837.0 &   15,840.0 &   22,484.5 &{\bf364,200.0}\\
           Asteroids &              14 &      871.3 &  36,517.3 &    1,458.7 &    1,193.2 &{\bf2,035.4}&    1,745.1 &      1,021.9 \\
            Atlantis &               4 &   13,463.0 &  26,575.0 &  292,491.0 &  319,688.0 &{\bf445,360.0}&  330,647.0 &    423,252.0 \\
          Bank Heist &              18 &       21.7 &     644.5 &      312.7 &      886.0 &{\bf1,129.3}&      876.6 &      1,004.6 \\
         Battle Zone &              18 &    3,560.0 &  33,030.0 &   23,750.0 &   24,740.0 &{\bf31,320.0}&   25,520.0 &     30,650.0 \\
          Beam Rider &               9 &      254.6 &  14,961.0 &    9,743.2 &   17,417.2 &   14,591.3 &   31,181.3 &{\bf37,412.2}\\
             Berzerk &              18 &      196.1 &   2,237.5 &      493.4 &    1,011.1 &      910.6 &      865.9 &{\bf2,178.6}\\
             Bowling &               6 &       35.2 &     146.5 &       56.5 &{\bf69.6}&       65.7 &       52.0 &         50.4 \\
              Boxing &              18 &       -1.5 &       9.6 &       70.3 &       73.5 &       77.3 &       72.3 &{\bf79.2}\\
            Breakout &               4 &        1.6 &      27.9 &      354.5 &      368.9 &{\bf411.6}&      343.0 &        354.6 \\
           Centipede &              18 &    1,925.5 &  10,321.9 &    3,973.9 &    3,853.5 &    4,881.0 &    3,489.1 &{\bf5,570.2}\\
     Chopper Command &              18 &      644.0 &   8,930.0 &    5,017.0 &    3,495.0 &    3,784.0 &    4,635.0 &{\bf8,058.0}\\
       Crazy Climber &               9 &    9,337.0 &  32,667.0 &   98,128.0 &  113,782.0 &  124,566.0 &  127,512.0 &{\bf127,853.0}\\
            Defender &              18 &    1,965.5 &  14,296.0 &   15,917.5 &   27,510.0 &   33,996.0 &   23,666.5 &{\bf34,415.0}\\
        Demon Attack &               6 &      208.3 &   3,442.8 &   12,550.7 &   69,803.4 &   56,322.8 &   61,277.5 &{\bf73,371.3}\\
         Double Dunk &              18 &      -16.0 &     -14.4 &       -6.0 &       -0.3 &       -0.8 &{\bf16.0}&        -10.7 \\
              Enduro &               9 &      -81.8 &     740.2 &      626.7 &    1,216.6 &    2,077.4 &    1,831.0 &{\bf2,223.9}\\
       Fishing Derby &              18 &      -77.1 &       5.1 &       -1.6 &        3.2 &       -4.1 &        9.8 &{\bf17.0}\\
             Freeway &               3 &        0.1 &      25.6 &       26.9 &       28.8 &        0.2 &{\bf28.9}&         28.2 \\
           Frostbite &              18 &       66.4 &   4,202.8 &      496.1 &    1,448.1 &    2,332.4 &    3,510.0 &{\bf4,038.4}\\
              Gopher &               8 &      250.0 &   2,311.0 &    8,190.4 &   15,253.0 &   20,051.4 &   34,858.8 &{\bf105,148.4}\\
            Gravitar &              18 &      245.5 &   3,116.0 &{\bf298.0}&      200.5 &      297.0 &      269.5 &        167.0 \\
            H.E.R.O. &              18 &    1,580.3 &  25,839.4 &   14,992.9 &   14,892.5 &   15,207.9 &{\bf20,889.9}&     15,459.2 \\
          Ice Hockey &              18 &       -9.7 &       0.5 &       -1.6 &       -2.5 &       -1.3 &       -0.2 &{\bf0.5}\\
          James Bond &              18 &       33.5 &     368.5 &      697.5 &      573.0 &      835.5 &{\bf3,961.0}&        585.0 \\
            Kangaroo &              18 &      100.0 &   2,739.0 &    4,496.0 &   11,204.0 &   10,334.0 &{\bf12,185.0}&        861.0 \\
               Krull &              18 &    1,151.9 &   2,109.1 &    6,206.0 &    6,796.1 &{\bf8,051.6}&    6,872.8 &      7,658.6 \\
      Kung-Fu Master &              14 &      304.0 &  20,786.8 &   20,882.0 &   30,207.0 &   24,288.0 &   31,676.0 &{\bf37,484.0}\\
 Montezuma's Revenge &              18 &       25.0 &   4,182.0 &       47.0 &       42.0 &       22.0 &{\bf51.0}&         24.0 \\
         Ms. Pac-Man &               9 &      197.8 &  15,375.0 &    1,092.3 &    1,241.3 &{\bf2,250.6}&    1,865.9 &      1,007.8 \\
      Name This Game &               6 &    1,747.8 &   6,796.0 &    6,738.8 &    8,960.3 &   11,185.1 &   10,497.6 &{\bf13,637.9}\\
             Phoenix &               8 &    1,134.4 &   6,686.2 &    7,484.8 &   12,366.5 &   20,410.5 &   16,903.6 &{\bf63,597.0}\\
            Pitfall! &              18 &     -348.8 &   5,998.9 &     -113.2 &     -186.7 &{\bf-46.9}&     -427.0 &       -243.6 \\
                Pong &               3 &      -18.0 &      15.5 &       18.0 &{\bf19.1}&       18.8 &       18.9 &         18.4 \\
         Private Eye &              18 &      662.8 &  64,169.1 &      207.9 &     -575.5 &      292.6 &      670.7 &{\bf1,277.6}\\
              Q*Bert &               6 &      183.0 &  12,085.0 &    9,271.5 &   11,020.8 &{\bf14,175.8}&    9,944.0 &     14,063.0 \\
          River Raid &              18 &      588.3 &  14,382.2 &    4,748.5 &   10,838.4 &{\bf16,569.4}&   11,807.2 &     16,496.8 \\
         Road Runner &              18 &      200.0 &   6,878.0 &   35,215.0 &   43,156.0 &{\bf58,549.0}&   52,264.0 &     54,630.0 \\
            Robotank &              18 &        2.4 &       8.9 &       58.7 &       59.1 &{\bf62.0}&       56.2 &         24.7 \\
            Seaquest &              18 &      215.5 &  40,425.8 &    4,216.7 &   14,498.0 &{\bf37,361.6}&   25,463.7 &      1,431.2 \\
              Skiing &               3 &  -15,287.4 &  -3,686.6 &  -12,142.1 &  -11,490.4 &  -11,928.0 &{\bf-10,169.1}&    -18,955.8 \\
             Solaris &              18 &    2,047.2 &  11,032.6 &    1,295.4 &      810.0 &    1,768.4 &{\bf2,272.8}&        280.6 \\
      Space Invaders &               6 &      182.6 &   1,464.9 &    1,293.8 &    2,628.7 &    5,993.1 &    3,912.1 &{\bf8,978.0}\\
         Star Gunner &              18 &      697.0 &   9,528.0 &   52,970.0 &   58,365.0 &   90,804.0 &   61,582.0 &{\bf127,073.0}\\
           Surround  &               5 &       -9.7 &       5.4 &       -6.0 &        1.9 &        4.0 &{\bf5.9}&         -0.2 \\
              Tennis &              18 &      -21.4 &      -6.7 &{\bf11.1}&       -7.8 &        4.4 &       -5.3 &        -13.2 \\
          Time Pilot &              10 &    3,273.0 &   5,650.0 &    4,786.0 &{\bf6,608.0}&    6,601.0 &    5,963.0 &      4,871.0 \\
           Tutankham &               8 &       12.7 &     138.3 &       45.6 &       92.2 &       48.0 &       56.9 &{\bf108.6}\\
         Up and Down &               6 &      707.2 &   9,896.1 &    8,038.5 &   19,086.9 &{\bf24,759.2}&   12,157.4 &     22,681.3 \\
             Venture &              18 &       18.0 &   1,039.0 &      136.0 &       21.0 &{\bf200.0}&       94.0 &         29.0 \\
       Video Pinball &               9 &   20,452.0 &  15,641.1 &  154,414.1 &  367,823.7 &  110,976.2 &  295,972.8 &{\bf447,408.6}\\
       Wizard Of Wor &              10 &      804.0 &   4,556.0 &    1,609.0 &    6,201.0 &    7,054.0 &    5,727.0 &{\bf10,471.0}\\
       Yars' Revenge &              18 &    1,476.9 &  47,135.2 &    4,577.5 &    6,270.6 &   25,976.5 &    4,687.4 &{\bf58,145.9}\\
              Zaxxon &              18 &      475.0 &   8,443.0 &    4,412.0 &    8,593.0 &   10,164.0 &    9,474.0 &{\bf11,320.0}\\
  \end{tabular}
\end{center}
\end{table}

\begin{table}[!htb]
\caption{Normalized scores across all games. Starting with {\bf 30 no-op} actions.}
\footnotesize
\begin{center}
\begin{tabular}{l|rrr|rr}
       {\sc Games} &   {\sc DQN} &       {\sc DDQN} &       {\sc Duel} &{\sc Prior.} & {\sc Prior. Duel.} \\
\hline
               Alien &    20.2\% &    51.0\% &{\bf61.4\%}&    57.6\% &      53.8\% \\
              Amidar &    56.7\% &   104.3\% &{\bf137.1\%}&   107.0\% &     133.7\% \\
             Assault &   781.0\% &   995.1\% &   846.5\% &  1433.7\% &{\bf2166.0\%}\\
             Asterix &    50.0\% &   206.8\% &   337.4\% &   377.6\% &{\bf4520.1\%}\\
           Asteroids &     1.4\% &     0.0\% &{\bf4.5\%}&     4.1\% &       1.0\% \\
            Atlantis &  1651.2\% &   576.1\% &  2285.3\% &  2129.3\% &{\bf2366.9\%}\\
          Bank Heist &    59.7\% &   137.6\% &{\bf216.2\%}&   140.8\% &     201.5\% \\
         Battle Zone &    79.1\% &    84.2\% &{\bf99.9\%}&    83.8\% &      95.2\% \\
          Beam Rider &    49.9\% &    81.0\% &    71.2\% &   139.0\% &{\bf180.6\%}\\
             Berzerk &    18.4\% &    44.0\% &    53.8\% &    47.2\% &{\bf131.1\%}\\
             Bowling &    19.8\% &{\bf32.7\%}&    30.8\% &    18.0\% &      17.1\% \\
              Boxing &   732.5\% &   762.1\% &{\bf827.1\%}&   795.5\% &     823.1\% \\
            Breakout &  1334.5\% &{\bf1449.2\%}&  1194.5\% &  1294.3\% &    1266.6\% \\
           Centipede &    25.9\% &    33.4\% &    55.1\% &    23.9\% &{\bf56.4\%}\\
     Chopper Command &    80.8\% &    76.0\% &   158.2\% &   118.4\% &{\bf188.1\%}\\
       Crazy Climber &   399.1\% &   425.2\% &   530.1\% &   520.5\% &{\bf604.6\%}\\
            Defender &   131.3\% &   205.3\% &{\bf248.8\%}&   179.7\% &     243.1\% \\
        Demon Attack &   659.6\% &  3182.8\% &  3335.0\% &  3941.6\% &{\bf3998.3\%}\\
         Double Dunk &   557.7\% &   607.9\% &   866.5\% &{\bf1723.3\%}&     280.5\% \\
              Enduro &    84.7\% &   140.8\% &   262.4\% &   243.2\% &{\bf268.0\%}\\
       Fishing Derby &   163.8\% &   202.4\% &{\bf260.7\%}&   247.7\% &     251.1\% \\
             Freeway &   104.0\% &   112.5\% &     0.1\% &{\bf114.0\%}&     111.3\% \\
           Frostbite &    17.1\% &    37.9\% &   107.9\% &   101.1\% &{\bf172.1\%}\\
              Gopher &   395.4\% &   676.7\% &   717.5\% &  1495.6\% &{\bf4831.3\%}\\
            Gravitar &     9.4\% &     7.5\% &{\bf13.1\%}&    11.8\% &       2.0\% \\
            H.E.R.O. &    65.1\% &    64.1\% &    66.4\% &{\bf73.9\%}&      67.1\% \\
          Ice Hockey &    76.9\% &    70.0\% &    96.4\% &{\bf103.2\%}&      89.6\% \\
          James Bond &   270.1\% &   485.4\% &   468.8\% &{\bf1869.8\%}&     286.0\% \\
            Kangaroo &   241.6\% &   433.8\% &   496.2\% &{\bf541.3\%}&      58.3\% \\
               Krull &   639.3\% &   592.3\% &{\bf923.1\%}&   761.6\% &     822.2\% \\
      Kung-Fu Master &   114.8\% &   131.0\% &   151.4\% &   174.9\% &{\bf214.1\%}\\
 Montezuma's Revenge &{\bf0.0\%}&     0.0\% &     0.0\% &     0.0\% &       0.0\% \\
         Ms. Pac-Man &    41.8\% &    36.2\% &    89.9\% &{\bf93.5\%}&      45.5\% \\
      Name This Game &   102.8\% &   144.6\% &   168.1\% &   173.3\% &{\bf230.7\%}\\
             Phoenix &   119.2\% &   177.3\% &   344.5\% &   281.3\% &{\bf1073.3\%}\\
            Pitfall! &    -0.8\% &     3.0\% &{\bf3.4\%}&    -1.9\% &       3.4\% \\
                Pong &   114.0\% &   117.8\% &{\bf118.2\%}&   117.1\% &     118.0\% \\
         Private Eye &     0.2\% &     0.2\% &     0.1\% &{\bf0.3\%}&       0.3\% \\
              Q*Bert &    97.5\% &   112.3\% &{\bf143.4\%}&   121.1\% &     139.9\% \\
          River Raid &    38.3\% &    85.8\% &{\bf125.6\%}&    83.6\% &     122.1\% \\
         Road Runner &   504.7\% &   563.2\% &{\bf887.4\%}&   735.3\% &     793.3\% \\
            Robotank &   631.5\% &   643.7\% &{\bf645.1\%}&   617.5\% &     259.5\% \\
            Seaquest &    13.8\% &    39.0\% &{\bf119.5\%}&    62.6\% &       2.1\% \\
              Skiing &    31.6\% &    63.3\% &{\bf64.6\%}&    55.6\% &     -22.3\% \\
             Solaris &    20.3\% &    16.5\% &     9.1\% &{\bf27.7\%}&      -9.9\% \\
      Space Invaders &   101.6\% &   156.3\% &   412.9\% &   178.7\% &{\bf997.2\%}\\
         Star Gunner &   559.3\% &   620.5\% &   924.0\% &   653.4\% &{\bf1298.3\%}\\
           Surround  &    26.5\% &    43.2\% &    86.9\% &{\bf114.6\%}&      67.6\% \\
              Tennis &{\bf231.3\%}&     6.8\% &   186.2\% &   153.2\% &     153.2\% \\
          Time Pilot &    78.4\% &   287.2\% &{\bf487.5\%}&   338.9\% &     239.9\% \\
           Tutankham &    36.3\% &   132.5\% &   128.1\% &   123.7\% &{\bf150.1\%}\\
         Up and Down &    84.7\% &   201.1\% &{\bf397.9\%}&   140.0\% &     298.8\% \\
             Venture &    13.7\% &     8.3\% &{\bf41.9\%}&     4.5\% &       4.0\% \\
       Video Pinball &  1113.7\% &  1754.3\% &   555.9\% &  1596.2\% &{\bf2712.2\%}\\
       Wizard Of Wor &    51.0\% &   165.2\% &   173.9\% &   101.1\% &{\bf281.1\%}\\
       Yars' Revenge &    29.1\% &    16.7\% &    90.4\% &    16.1\% &{\bf129.2\%}\\
              Zaxxon &    58.3\% &   110.8\% &   141.3\% &   114.2\% &{\bf151.6\%}\\
\end{tabular}
\end{center}
\end{table}

\begin{table}[t]
\caption{Normalized scores across all games. Starting with {\bf Human Starts}.}
\footnotesize
\begin{center}
\begin{tabular}{l|rrr|rr}
              {\sc Games} &   {\sc DQN} &       {\sc DDQN} &       {\sc Duel} &{\sc Prior.} & {\sc Prior. Duel.} \\
\hline 
               Alien &     8.1\% &    14.5\% &{\bf21.8\%}&    19.3\% &      11.1\% \\
              Amidar &    10.9\% &    10.3\% &    10.5\% &     7.7\% &{\bf14.8\%}\\
             Assault &   719.2\% &  1275.9\% &   828.6\% &  1381.5\% &{\bf2334.4\%}\\
             Asterix &    40.8\% &   226.2\% &   212.7\% &   302.8\% &{\bf4938.4\%}\\
           Asteroids &     1.6\% &     0.9\% &{\bf3.3\%}&     2.5\% &       0.4\% \\
            Atlantis &  2128.0\% &  2335.5\% &{\bf3293.9\%}&  2419.0\% &    3125.3\% \\
          Bank Heist &    46.7\% &   138.8\% &{\bf177.8\%}&   137.3\% &     157.8\% \\
         Battle Zone &    68.5\% &    71.9\% &{\bf94.2\%}&    74.5\% &      91.9\% \\
          Beam Rider &    64.5\% &   116.7\% &    97.5\% &   210.3\% &{\bf252.7\%}\\
             Berzerk &    14.6\% &    39.9\% &    35.0\% &    32.8\% &{\bf97.1\%}\\
             Bowling &    19.2\% &{\bf30.9\%}&    27.5\% &    15.1\% &      13.7\% \\
              Boxing &   648.2\% &   677.0\% &   711.2\% &   666.7\% &{\bf728.5\%}\\
            Breakout &  1341.9\% &  1396.7\% &{\bf1559.0\%}&  1298.3\% &    1342.4\% \\
           Centipede &    24.4\% &    23.0\% &    35.2\% &    18.6\% &{\bf43.4\%}\\
     Chopper Command &    52.8\% &    34.4\% &    37.9\% &    48.2\% &{\bf89.5\%}\\
       Crazy Climber &   380.6\% &   447.7\% &   493.9\% &   506.5\% &{\bf508.0\%}\\
            Defender &   113.2\% &   207.2\% &   259.8\% &   176.0\% &{\bf263.2\%}\\
        Demon Attack &   381.6\% &  2151.6\% &  1734.8\% &  1888.0\% &{\bf2261.9\%}\\
         Double Dunk &   622.5\% &   982.5\% &   948.7\% &{\bf1998.7\%}&     328.7\% \\
              Enduro &    86.2\% &   158.0\% &   262.7\% &   232.7\% &{\bf280.5\%}\\
       Fishing Derby &    91.8\% &    97.7\% &    88.8\% &   105.7\% &{\bf114.5\%}\\
             Freeway &   105.1\% &   112.5\% &     0.6\% &{\bf113.1\%}&     110.2\% \\
           Frostbite &    10.4\% &    33.4\% &    54.8\% &    83.2\% &{\bf96.0\%}\\
              Gopher &   385.3\% &   727.9\% &   960.8\% &  1679.2\% &{\bf5089.7\%}\\
            Gravitar &{\bf1.8\%}&    -1.6\% &     1.8\% &     0.8\% &      -2.7\% \\
            H.E.R.O. &    55.3\% &    54.9\% &    56.2\% &{\bf79.6\%}&      57.2\% \\
          Ice Hockey &    79.2\% &    70.8\% &    82.4\% &    92.5\% &{\bf99.4\%}\\
          James Bond &   198.2\% &   161.0\% &   239.4\% &{\bf1172.4\%}&     164.6\% \\
            Kangaroo &   166.6\% &   420.8\% &   387.8\% &{\bf457.9\%}&      28.8\% \\
               Krull &   528.0\% &   589.7\% &{\bf720.8\%}&   597.7\% &     679.8\% \\
      Kung-Fu Master &   100.5\% &   146.0\% &   117.1\% &   153.2\% &{\bf181.5\%}\\
 Montezuma's Revenge &     0.5\% &     0.4\% &    -0.1\% &{\bf0.6\%}&      -0.0\% \\
         Ms. Pac-Man &     5.9\% &     6.9\% &{\bf13.5\%}&    11.0\% &       5.3\% \\
      Name This Game &    98.9\% &   142.9\% &   186.9\% &   173.3\% &{\bf235.5\%}\\
             Phoenix &   114.4\% &   202.3\% &   347.2\% &   284.0\% &{\bf1125.1\%}\\
            Pitfall! &     3.7\% &     2.6\% &{\bf4.8\%}&    -1.2\% &       1.7\% \\
                Pong &   107.6\% &{\bf110.9\%}&   110.0\% &   110.1\% &     108.6\% \\
         Private Eye &    -0.7\% &    -1.9\% &    -0.6\% &     0.0\% &{\bf1.0\%}\\
              Q*Bert &    76.4\% &    91.1\% &{\bf117.6\%}&    82.0\% &     116.6\% \\
          River Raid &    30.2\% &    74.3\% &{\bf115.9\%}&    81.3\% &     115.3\% \\
         Road Runner &   524.3\% &   643.2\% &{\bf873.7\%}&   779.6\% &     815.1\% \\
            Robotank &   863.3\% &   868.7\% &{\bf913.3\%}&   824.4\% &     341.1\% \\
            Seaquest &    10.0\% &    35.5\% &{\bf92.4\%}&    62.8\% &       3.0\% \\
              Skiing &    27.1\% &    32.7\% &    29.0\% &{\bf44.1\%}&     -31.6\% \\
             Solaris &    -8.4\% &   -13.8\% &    -3.1\% &{\bf2.5\%}&     -19.7\% \\
      Space Invaders &    86.7\% &   190.8\% &   453.1\% &   290.8\% &{\bf685.9\%}\\
         Star Gunner &   591.9\% &   653.0\% &  1020.3\% &   689.4\% &{\bf1431.0\%}\\
           Surround  &    24.7\% &    76.9\% &    91.1\% &{\bf103.2\%}&      62.9\% \\
              Tennis &{\bf220.8\%}&    92.1\% &   175.0\% &   109.6\% &      55.6\% \\
          Time Pilot &    63.7\% &{\bf140.3\%}&   140.0\% &   113.2\% &      67.2\% \\
           Tutankham &    26.2\% &    63.3\% &    28.1\% &    35.2\% &{\bf76.4\%}\\
         Up and Down &    79.8\% &   200.0\% &{\bf261.8\%}&   124.6\% &     239.1\% \\
             Venture &    11.6\% &     0.3\% &{\bf17.8\%}&     7.4\% &       1.1\% \\
       Video Pinball &   987.2\% &  2351.6\% &   709.5\% &  1892.3\% &{\bf2860.5\%}\\
       Wizard Of Wor &    21.5\% &   143.8\% &   166.6\% &   131.2\% &{\bf257.6\%}\\
       Yars' Revenge &     6.8\% &    10.5\% &    53.7\% &     7.0\% &{\bf124.1\%}\\
              Zaxxon &    49.4\% &   101.9\% &   121.6\% &   112.9\% &{\bf136.1\%}\\
            \hline
            {\bf Mean}   &    219.6\% & 332.9\% &  343.8\% & 386.7\% & {\bf 567.0}\%\\
            {\bf Median}   &68.5\% & 110.9\% & {\bf117.1}\% & 112.9\% & 115.3\%\\
\end{tabular}
\end{center}
\end{table}

% %\newpage
% \section{Score Tables}
% \input{small_appendix.tex}

\end{document}